\documentclass{article}

\usepackage[preprint]{neurips_2024}

\usepackage[utf8]{inputenc} 
\usepackage[T1]{fontenc}    
\usepackage{hyperref}       
\usepackage{url}            
\usepackage{booktabs}       
\usepackage{amsfonts}       
\usepackage{nicefrac}       
\usepackage{microtype}      
\usepackage{xcolor}         
\usepackage{graphicx}
\usepackage{amsmath}
\usepackage{tikz}
\usepackage{multirow}
\usepackage{adjustbox}
\usepackage{float}
\usepackage{placeins}

\setcitestyle{numbers,square,commas}

\title{Detecting Transportation Mode Using Dense Smartphone GPS Trajectories and Transformer Models}

\author{
  Yuandong Zhang\thanks{These authors contributed equally to this work.} \\
  University of California\\
  San Diego, CA, USA \\
  \And
  Othmane Echchabi\footnotemark[1]\\
  McGill University \\
  Mila - Quebec AI Institute \\
  Montréal, QC, Canada
  \AND
  Tianshu Feng \\
  University of Pennsylvania \\
  Philadelphia, PA, USA \\
  \And
  Wenyi Zhang \\
  Duke Kunshan University \\
  Kunshan, Jiangsu, China \\
  \And
  Hsuai-Kai Liao \\
  Duke Kunshan University \\
  Kunshan, Jiangsu, China \\
  \And
  Charles Chang\thanks{Corresponding author. Email: charles.c.chang@dukekunshan.edu.cn}  \\
  Duke Kunshan University \\
  Kunshan, Jiangsu, China \\
}

\begin{document}

\maketitle

\begin{abstract}
Transportation mode detection is an important topic within GeoAI and transportation research. In this study, we introduce \textsc{SpeedTransformer}, a novel Transformer-based model that relies solely on speed inputs to infer transportation modes from dense smartphone GPS trajectories. In benchmark experiments, \textsc{SpeedTransformer} outperformed traditional deep learning models, such as the Long Short-Term Memory (LSTM) network. Moreover, the model demonstrated strong flexibility in transfer learning, achieving high accuracy across geographical regions after fine-tuning with small datasets. Finally, we deployed the model in a real-world experiment, where it consistently outperformed baseline models under complex built environments and high data uncertainty. These findings suggest that Transformer architectures, when combined with dense GPS trajectories, hold substantial potential for advancing transportation mode detection and broader mobility-related research.

\end{abstract}

\section{Introduction}

The study of human mobility patterns—how individuals move across space—has become an important topic across geography, transportation science, public health, and climate change science \citep{gonzalez2008understanding, schuessler2009processing, yao2020understanding, shaw2016human, bonaccorsi2020economic, mcmichael2020human, barbosa2021uncovering, zook2015geographies, guo2015mobile, tao2018analytics}. One key aspect of human mobility is transportation mode choice, the accurate estimation of which is essential for understanding individual carbon emissions and associated health benefits \citep{girod2013influence, tajalli2017relationships}.

Traditionally, transportation surveys were used to estimate individual's choices of transportation mode \citep{wang2015carbon}. However, two major technological advancements in recent years have reshaped transportation mode detection. First, smartphones—equipped with GPS, accelerometers, gyroscopes, and cellular network connectivity—have enabled the creation of detailed mobility datasets derived from mobile applications \citep{molloy_mobis_2023}, social media platforms \citep{Preotiuc-Pietro2013}, travel cards \citep{Gordon2013AutomatedIO}, and cellphone signals \citep{lu2012predictability}. These datasets surpassed traditional transportation surveys in dimensionality, accuracy, variety, and volume \citep{Barbosa2015, goodchild2013quality}.

Second, advancements in machine learning (ML) have greatly improved the ability to extract meaningful information from mobility datasets. Machine learning techniques—from ensemble methods to deep learning models—now exploit increasingly fine-grained spatiotemporal data, including human mobility traces, substantially improving the accuracy of transportation-mode prediction \citep{Pappalardo2023}. More recently, Transformer \citep{vaswani_attention_2017} architectures are adopted for human mobility research given their self-attention mechanisms, which can help capture nonlinear spatiotemporal dependencies and high-order motion features. Furthermore, their capacity for inductive transfer enables cross-regional generalization, potentially allowing models to adapt to diverse transportation infrastructures using minimal labeled data.

Nevertheless, significant challenges remained. Smartphone-derived mobility datasets, while highly granular, often exhibited inconsistent quality due to the absence of standardized data collection protocols and inaccuracies in geographic information, complicating modeling process. Model performance frequently depended on derived features such as acceleration, while complex preprocessing tasks—such as geocoding and geotagging—further increased data uncertainty. Moreover, extensive feature engineering used to aggregate raw GPS trajectories often resulted in the loss of critical sequential information essential for accurate mobility modeling \citep{jahangiri_applying_2015}.

Privacy posed another major challenge. Mobility applications commonly collected sensitive information, including geographic locations, trip details (e.g., start and end times), and personal or device identifiers. For example, \citet{xu2017trajectory} demonstrated that even anonymized mobility data could be re-identified with 73–91\% accuracy, underscoring the difficulty of ensuring privacy protection. Such risks frequently discouraged individuals from sharing their mobility data \citep{Rzeszewski2018}, constraining data availability.

Moreover, mobility models have often lacked generalizability across geographic regions, particularly when trained on isolated or geographically homogeneous datasets. Human mobility behaviors vary widely across countries and regions due to differences in road infrastructure, speed regulations, and cultural norms. Nevertheless, most studies have assessed model performance using train–test splits from the same dataset \citep{Zhao2016}, which fail to capture real-world adaptability. Models fine-tuned on benchmarks such as Geolife \citep{zheng_Geolife_2010} frequently required substantial recalibration when applied to other contexts. The proliferation of deep learning frameworks, hyperparameter tuning strategies, and architectural variations has further complicated cross-regional reproducibility in mobility modeling research \citep{graser_mobilitydl_2024}.

Finally, existing mobility models have often struggled to perform reliably under real-world conditions, which are considerably more unpredictable and complex than curated research datasets suggest. Everyday travel involves nuances such as short trips and unexpected errands that traditional models frequently misclassified \citep{Pappalardo2023}. Moreover, GPS signal quality varied across smartphone models and was highly sensitive to environmental factors, including urban canyons and signal obstructions \citep{Zandbergen2009, Cui2001}. These real-world data inconsistencies—typically underrepresented in benchmark datasets—underscore the need to validate mobility models under realistic conditions that reflect the full variability of human movement.

To address these challenges, we introduced a Transformer-based deep learning model that uses a simple input—instantaneous speed—to achieve highly accurate transportation mode detection. We refer to this model as \textsc{SpeedTransformer}. Using transportation mode classification as a case study, we demonstrate the model’s capability and versatility. Our contributions are threefold. First, we show that a Transformer-based neural network can achieve state-of-the-art performance using only speed as input. By leveraging positional encoding and multi-head attention, our model captured complex temporal patterns without elaborate feature engineering, thereby mitigating both privacy concerns and computational demands. Second, we demonstrate robust cross-regional generalizability through transfer learning: a model pre-trained on data primarily collected in Switzerland maintained exceptional performance when fine-tuned on samples from Beijing. Finally, we validated our approach under real-world conditions by developing a novel smartphone mini-program and recruiting 348 participants for a one-month field experiment. The experimental results confirmed that our model’s advantages translated effectively from controlled environments to practical applications characterized by real-world uncertainty and variability.

\section{Related Work}
\smallskip
\noindent
Transportation mode detection represents a central dimension of human mobility research, with downstream applications ranging from carbon footprint estimation \citep{Manzoni2010} and tourism recommendations \citep{Xiao2025} to traffic management \citep{Prelipcean2016} and public health interventions \citep{Aleta2022}. With the proliferation of GPS-enabled devices and advances in machine learning, the field has undergone substantial progress in recent years, moving beyond traditional social scientific methods such as transportation surveys and adopting data-driven methodologies grounded in machine learning. Notable developments have emerged in both classical machine learning algorithms and deep learning approaches.
 
\subsection{Machine Learning for Transportation Mode Detection and its Challenges}

Classical machine learning (ML) algorithms formed the foundation of early data-driven approaches to transportation mode detection. These methods typically converted sequential GPS trajectories into statistical, tabular representations, with notable implementations including Decision Trees \citep{zheng_understanding_2008}, Random Forests \citep{stenneth_transportation_2011, jahangiri_applying_2015}, and Support Vector Machines \citep{bolbol_inferring_2012}. \citet{stenneth_transportation_2011} demonstrated the effectiveness of these models through systematic evaluation, while \citet{jahangiri_applying_2015} highlighted their versatility across different transportation modes, variations, and contexts. The logic underlying classical ML algorithms was conceptually similar to that of rule-based models, which relied on indicative features computed from GPS trajectories and used them to infer travel modes statistically. For example, mobility features derived from dense GPS trajectories—such as average speed and acceleration rate—were often effective in distinguishing between driving, walking, and cycling.  

Although computationally efficient, these approaches depended heavily on domain expertise for feature engineering and performed poorly when handling variable-length inputs. In many inner-city trips, for instance, driving was only marginally faster than cycling, rendering average speed an unreliable discriminator. More advanced statistical features could improve performance but required specialized expertise and local contextual knowledge that were seldom available and sometimes unreliable.

Moreover, such methods posed risks to privacy and anonymity, as even a small number of precise and longitudinal GPS points could be sufficient to re-identify individuals \citep{de2013unique}. Growing concerns regarding the collection, management, and disclosure of personal GPS data, together with advances in re-identification techniques, have further raised ethical issues surrounding GPS trajectory research \citep{michael2006emerging, klasnja2009exploring, minch2004privacy}. In response, scholars and regulators have increasingly advocated for privacy-preserving techniques in mobility research—including applications such as transportation mode detection—as alternatives to classical machine learning approaches that depend on rich location features and extensive feature engineering \citep{ng2002price, krumm2009survey, shin2012privacy, thompson2022twelve, jiang2021location}.

Recurrent neural networks (RNNs), particularly Long Short-Term Memory (LSTM) networks \citep{hochreiter_long_1997}, emerged as a dominant approach for transportation mode prediction due to their capacity to capture temporal dependencies in sequential data. \citet{jiang_trajectorynet_2017} pioneered the application of LSTMs for mobility analysis, and \citet{asci_novel_2019} extended this work by incorporating attention mechanisms that improved performance across varied trip lengths. Hybrid architectures combining LSTMs with other neural components further advanced model accuracy—for example, the Conv-LSTM model proposed by \citet{nawaz_convolutional_2020} leveraged convolutional layers for spatial feature extraction prior to temporal processing. Other innovations included Convolutional Neural Networks (CNNs) \citep{dabiri_inferring_2018}, autoregressive flow models \citep{dutta_inferencing_2023}, and semi-supervised learning approaches \citep{dabiri_semi-supervised_2020}. Despite these accuracy gains, such sophisticated architectures often required substantial computational resources and complex pre-processing to structure data according to model specifications, thereby limiting their practicality for real-world deployment \citep{graser_mobilitydl_2024}.

\subsection{Transformers for Mobility Modeling}
Transformer architectures revolutionized sequence modeling through their reliance on self-attention and multi-head attention mechanisms \citep{cheng2016long, vaswani_attention_2017}. Models such as GPT-2 \citep{GPT2} and BERT \citep{BERT} demonstrated exceptional performance on sequence-based data, significantly surpassing classical machine learning models in capturing complex sequential patterns. This success motivated researchers to explore whether Transformers’ superior capacity for modeling long-range dependencies could similarly advance mobility modeling, where understanding the relationships between distant points within a trajectory is critical \citep{XueLLM}.

Recent applications of Transformer architectures to mobility studies have shown promise, although the field remains in its early stages. For example, \citet{hong_how_2022} incorporated Transformer components for next-location prediction while treating mode identification as a secondary task. \citet{liang_trajformer_2022} addressed challenges such as irregular spatiotemporal intervals but focused primarily on spatio-temporal dependencies rather than transportation mode prediction. More recently, \citet{ribeiro_deep_2024} achieved strong results using a vision Transformer approach \citep{dosovitskiy_image_2021}, although their implementation required converting trajectories into image-based representations. Similarly, \citet{drosouli2023tmd} converted GPS trajectory points into word tokens to apply the original BERT model \citep{BERT}, achieving notable results; however, this approach imposed a linguistic abstraction onto spatial data, making it less suitable for general-purpose mobility modeling.

Despite these advances, existing Transformer-based approaches for mobility modeling typically required extensive pre-processing, multiple input features, or auxiliary contextual information—making them computationally demanding and often impractical for real-world applications where only basic GPS data are available. Moreover, their generalizability across different geographical contexts has remained largely unexplored, as most evaluations have focused on performance within a single dataset rather than testing transferability between regions with distinct transportation infrastructures and mobility behaviors. 

\section{SpeedTransformer Architecture}
\label{sec:arch}

\begin{figure}[ht]
    \centering
    \includegraphics[width=\linewidth]{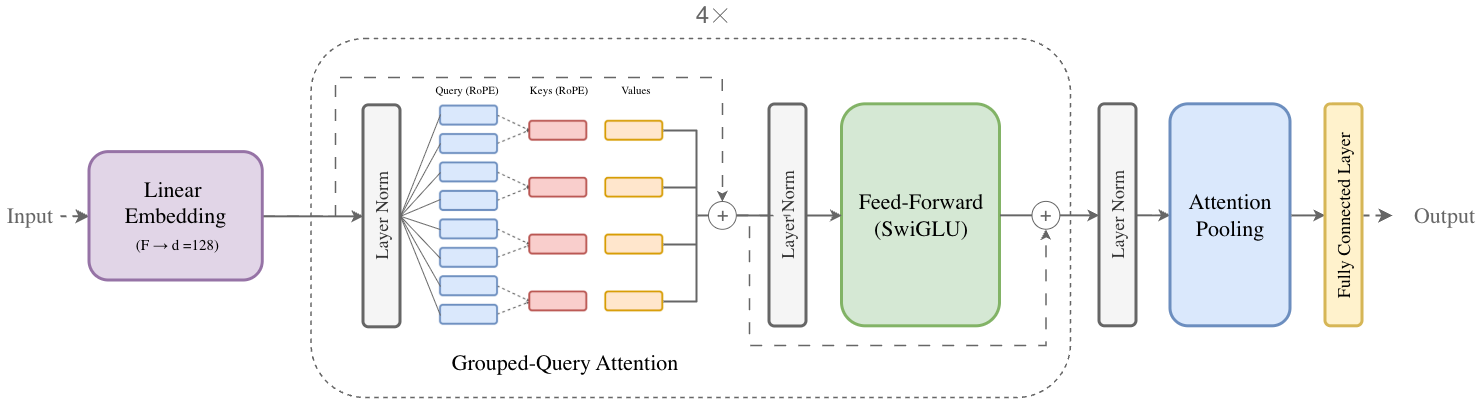}
    \caption{Transformer architecture}
    \label{fig:transformer_architecture}
\end{figure}
 
Our \textsc{SpeedTransformer} architecture adapts the Transformer encoder framework proposed by \citet{vaswani_attention_2017}, incorporating several key modifications to the data input, model structure, and output. First, the input sequences consist of instantaneous speeds computed from dense GPS trajectories collected via transportation applications on smartphones. These speed sequences are sampled at high frequencies—typically representing distances traveled over five to ten seconds—and therefore implicitly encoded higher-order motion features such as acceleration (the first derivative of speed) and jerk (the second derivative of speed). Given a complete speed sequence from trip start to end, each speed value serves as a token, which is subsequently embedded and transformed into a query vector representing its relationship to all other speed positions within the sequence.

Figure~\ref{fig:transformer_architecture} illustrates the overall model architecture. Our model requires only raw scalar speed sequences as trajectory input. To accommodate variable input lengths, each trajectory is segmented into fixed-length sequences of \( T = 200 \) consecutive speed samples using a sliding window with a stride of 50 (see Appendix~\ref{window_size} for further discussion on window size selection). This segmentation is applied directly to the original speed sequences without temporal resampling, allowing the model to remain robust to differences in sampling frequency across datasets.\footnote{For the details of temporal sampling frequency, see Appendix~\ref{appendix:temporal}.} Shorter sequences are zero-padded, and a key-padding mask is applied to ensure that padded tokens are ignored during attention and pooling.

Each scalar speed value \( s_t \) is linearly projected into a \( d = 128 \)-dimensional embedding space (see Appendix~\ref{appendix:embedding} for details on the embedding process). The sequence of embedded speed vectors is then processed by a modified Transformer encoder. Owing to its attention mechanism, the model is able to extract sequential dependencies—such as acceleration and jerk—from these speed embeddings, which is critical for differentiating transportation modes. When trained on sufficiently large empirical datasets, the model effectively optimizes its capacity to detect transportation modes.

We replace the standard sinusoidal positional encoding with Rotary Positional Embeddings (RoPE), which is applied directly to the query (\( Q \)) and key (\( K \)) vectors in the attention mechanism \citep{Su2021RoPE}. RoPE encodes positional information through position-dependent rotations indexed by the sequence order, enabling the attention mechanism to model relative temporal dependencies in a continuous and rotation-invariant manner. This formulation is particularly well-suited for sequential mobility signals such as speed trajectories, where relative temporal structure is more informative than absolute position.

The encoder consists of \( L = 4 \) Pre-Norm Transformer blocks. Each block contains two key components: a Grouped-Query Attention (GQA) layer \citep{Ainslie2023GQA}, which efficiently computes attention by grouping multiple query vectors per key–value pair, and a SwiGLU-activated feed-forward sublayer \citep{Shazeer2020GLU} that introduces non-linear transformations with improved gradient flow and expressivity (see Appendix~\ref{swiglu} for a detailed explanation of SwiGLU). The computation within each block is expressed as follows:

\begin{align}
\mathbf{z}_1 &= \mathbf{x} + \text{Dropout}(\text{GQA}(\text{LayerNorm}(\mathbf{x}))), \label{eq:gqa_block} \\
\mathbf{z}_2 &= \mathbf{z}_1 + \text{Dropout}(\text{FFN}_{\text{SwiGLU}}(\text{LayerNorm}(\mathbf{z}_1))), \label{eq:swiglu_block}
\end{align}

Equation~\ref{eq:gqa_block} describes the attention sub-layer, in which the input representation \( \mathbf{x} \in \mathbb{R}^{T \times d} \) (a sequence of \( T \) tokens, each of dimension \( d \)) is first normalized using layer normalization before being passed to the GQA mechanism. 

GQA divides the \(\mathbf{h}\) query heads into \(\mathbf{h_{kv}}\) groups, each group sharing a single key–value pair. This interpolates between multi‑head and multi‑query attention; using an intermediate number of groups retains much of the quality of multi‑head attention while reducing memory and compute cost. We set \(\mathbf{h} = 8\) query heads and \(\mathbf{h_{kv}} = 4\) key/value heads in our experiments.

Equation~\ref{eq:swiglu_block} represents the feed-forward transformation that followed the attention mechanism. The intermediate output \( \mathbf{z}_1 \) is normalized again and passed through a SwiGLU-activated feed-forward network, \( \text{FFN}_{\text{SwiGLU}}(\cdot) \), which introduces non-linear transformations to enhance representational capacity while maintaining computational efficiency. 

Each layer applies layer normalization before both the attention and the feed‑forward sub‑layers and uses dropout on the sub‑layer outputs before adding them back to the residual input, following the “residual dropout” strategy of the original Transformer

We employ \( h = 8 \) query heads and \( h_{kv} = 4 \) shared key/value heads, following the formulation of \citet{Ainslie2023GQA}. This design reduces both memory usage and computational cost by allowing multiple query heads to share the same key and value projections while preserving diversity across query subspaces. RoPE is applied to \( Q \) and \( K \) before computing scaled dot-product attention, and padding masks are introduced to ensure that zero-padded tokens did not contribute to the attention weights.

After the encoder stack, a final layer normalization is applied, producing the sequence representation $\mathbf{Z} = [\mathbf{z}_1, \dots, \mathbf{z}_T]$. The contextualized sequence is then aggregated through attention pooling. Specifically, each timestep embedding $\mathbf{z}_t$ receives a learnable scalar attention score $e_t$, which is normalized using a masked softmax function across valid time-steps to obtain weights $\alpha_t$:
\begin{align}
e_t &= \mathbf{w}_a^\top \mathbf{z}_t + b_a, \label{eq:att_score} \\
\alpha_t &= \frac{\exp(e_t) \cdot M_t}{\sum_{\tau=1}^T \exp(e_{\tau}) \cdot M_\tau}, \label{eq:att_weight}
\end{align}
\noindent where $\mathbf{w}_a$ and $b_a$ are learnable parameters, and $M_t \in \{0, 1\}$ is the mask for padding tokens. The pooled sequence embedding $\mathbf{c}$ is computed as the weighted sum $\mathbf{c} = \sum_{t=1}^T \alpha_t \mathbf{z}_t$. Finally, after dropout regularization, $\mathbf{c}$ is passed through a linear projection layer to produce the probability distribution $\hat{\mathbf{y}}$:
\begin{equation}
\hat{\mathbf{y}} = \text{softmax}(\mathbf{W}_\mathbf{c} + \mathbf{b}_c),
\label{eq:classifier}
\end{equation}
\noindent where $\mathbf{W}_c$ and $\mathbf{b}_c$ represent the classifier weights and bias.

\section{Datasets}
\begin{figure}[ht]
    \centering
    \includegraphics[width=0.7\textwidth]{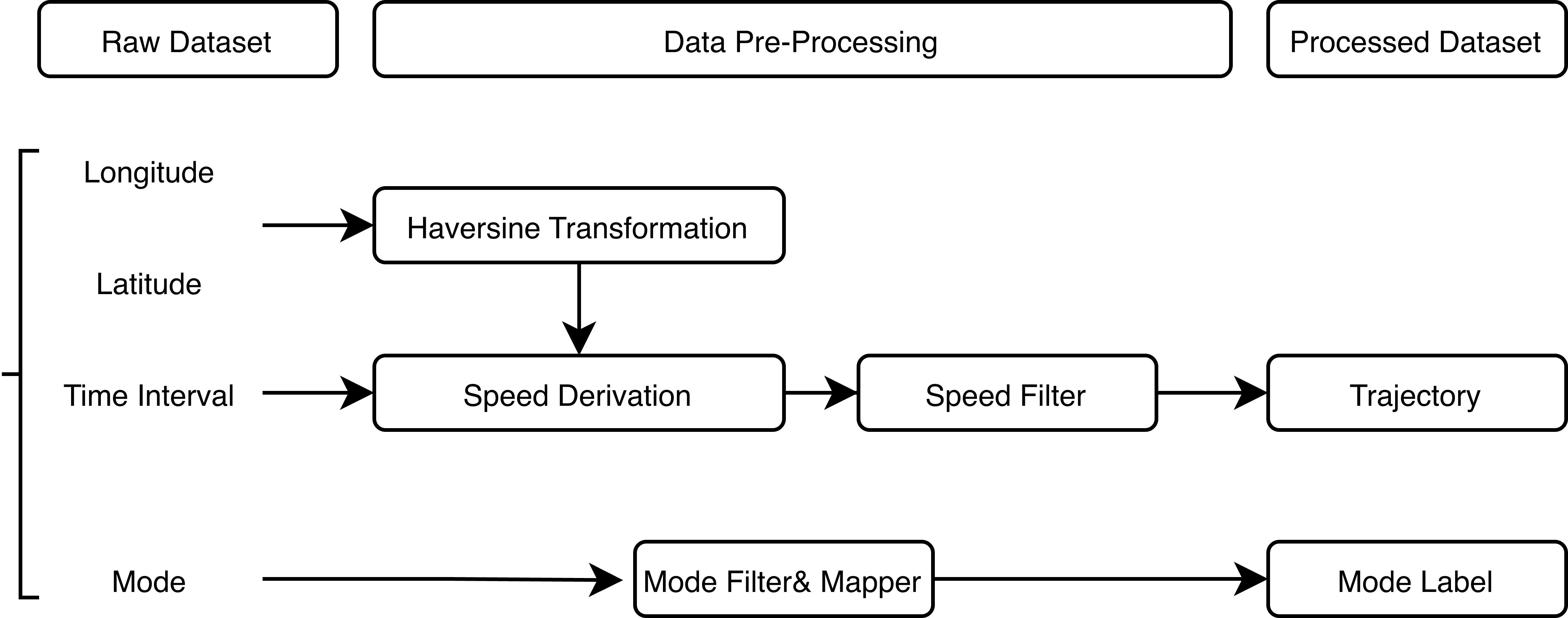}
    \caption{Data Pre-Processing}
    \label{fig:data_pre_processing}
\end{figure}
\begin{table}[b]
\centering
\caption{Comparison of MOBIS and Geolife Datasets}
\label{tab:dataset_comparison}
\begin{tabular}{@{}lrrrr@{}}
\toprule
\multirow{2}{*}{\textbf{Mode}} & \multicolumn{2}{c}{\textbf{Data Points}} & \multicolumn{2}{c}{\textbf{Unique Trips}} \\
\cmidrule(lr){2-3} \cmidrule(lr){4-5}
& \textbf{MOBIS} & \textbf{Geolife} & \textbf{MOBIS} & \textbf{Geolife} \\
\midrule
Bike & 4{,}251{,}028 & 746{,}098 & 40{,}171 & 1{,}555 \\
Bus & 4{,}606{,}409 & 1{,}061{,}196 & 74{,}324 & 1{,}847 \\
Car & 88{,}965{,}473 & 627{,}047 & 542{,}078 & 1{,}293 \\
Train & 8{,}379{,}962 & 788{,}250 & 130{,}964 & 772 \\
Walk & 38{,}463{,}266 & 1{,}215{,}054 & 743{,}425 & 3{,}960 \\
\midrule
\textbf{Total} & 144{,}666{,}138 & 4{,}437{,}645 & 1{,}530{,}962 & 9{,}427 \\
\bottomrule
\end{tabular}
\end{table}

We utilized two longitudinal tracking datasets—the widely adopted Geolife dataset \citep{zheng_Geolife_2010} and the Swiss MOBIS dataset \citep{heimgartner2024modal}—which offered complementary strengths for evaluating our model. Summary statistics are presented in Table~\ref{tab:dataset_comparison}. As illustrated in Figure~\ref{fig:data_pre_processing}, we standardized their data structures through a unified pre-processing pipeline: we consolidated transportation modes into five consistent categories (Bike, Bus, Car, Train, Walk), unified the trajectory input format across datasets, excluded multi-model trajectories, converted geographic coordinates into speed sequences, removed abnormal trips with erroneous location data, and standardized transportation mode labels. Detailed descriptions of the pre-processing procedures and raw dataset statistics are provided in Appendix~\ref{pre-processing}.

\subsection{MOBIS Dataset}
The MOBIS dataset \citep{molloy_mobis_2023} was derived from an eight-week randomized controlled trial (RCT) on transport pricing involving 3{,}680 participants in Switzerland. Each participant used the \textit{Catch-my-Day} mobile application (available for both iOS and Android), which continuously recorded GPS data via the device’s location services. The application captured daily travel patterns, storing raw trajectory data locally before uploading them to the MotionTag analytics platform, where trip segmentation and transportation mode inference were performed. This extensive data collection process produced 255.3 million GPS records and 1.58 million labeled trips in the raw dataset. After applying our standardized pre-processing pipeline described in Appendix~\ref{pre-processing}, the resulting MOBIS dataset contained 144.7 million data points across 1.53 million unique trips distributed among five transportation modes, as summarized in Table~\ref{tab:dataset_comparison}.

\subsection{Geolife Dataset}
Geolife \citep{zheng_Geolife_2010} was a widely used benchmark in transportation mode research, collected by Microsoft Research Asia from 182 users in Beijing over a five-year period (2007–2012). The dataset captured urban mobility through 17{,}000 trajectories covering approximately 1.2 million kilometers across Beijing’s complex transportation network. Data were recorded using various GPS loggers and GPS-enabled phones with different sampling rates, with 91\% of trajectories collected at high density (every 1–5 seconds or every 5–10 meters). In addition to routine commutes, the Geolife dataset included leisure and sports activities such as shopping, sightseeing, and cycling, offering rich contextual diversity in trip purposes. After pre-processing to align with the MOBIS data structure, the Geolife dataset contained 4.44 million data points and 9{,}427 unique trips—substantially smaller than MOBIS, yet providing valuable geographical and temporal diversity for evaluating our model’s performance.

\section{Experiments}
We evaluated \textsc{SpeedTransformer} under three experimental conditions. First, we benchmarked it against state-of-the-art transportation mode identification models on Geolife \citep{zheng_Geolife_2010}, enabling direct comparison with existing approaches. Second, we examined performance consistency across different geographical contexts by comparing it with classical LSTM baseline models on both the Swiss (MOBIS) \citep{heimgartner2024modal} and Chinese (Geolife) datasets. Finally, we assessed cross-regional transferability through fine-tuning experiments, in which models pretrained on Swiss data were adapted to Chinese mobility patterns using small sample subsets.

\subsection{Benchmarking Performance}
To evaluate SpeedTransformer's ability to achieve high accuracy with minimal input, we benchmarked against several state-of-the-art transportation mode identification models:
\begin{itemize}
\item \textbf{LSTM-Attention (Baseline)}: Our reconstructed baseline model implementing a classical bidirectional LSTM with attention mechanism \citep{hochreiter_long_1997}. This baseline is specifically designed to test whether pure attention-based mechanisms outperform recurrent networks augmented with attention, while maintaining the same minimal input requirement (speed only).
\item \textbf{Deep-ViT} \citep{ribeiro_deep_2024}: A Vision Transformer approach that transforms GPS features (speed, acceleration, bearing) into image representations using DeepInsight methodology before processing with a Vision Transformer architecture, combining traditional feature engineering with advanced deep learning.
\item \textbf{CE-RCRF} \citep{Zeng2023}: A sequence-to-sequence framework (TaaS) that processes entire trajectories using a Convolutional Encoder to extract high-level features and a Recurrent Conditional Random Field to maintain contextual information at both feature and label levels, with specialized bus-related features to distinguish high-speed modes.
\item \textbf{LSTM-based DNN} \citep{yu_semi-supervised_2020}: An ensemble of four LSTM networks that incorporates both time-domain trajectory attributes and frequency-domain statistics developed through discrete Fourier and wavelet transforms, creating a semi-supervised deep learning approach.
\item \textbf{SECA} \citep{dabiri_semi-supervised_2020}: A Semi-supervised Convolutional Autoencoder that integrates a convolutional-deconvolutional autoencoder with a CNN classifier to simultaneously leverage labeled and unlabeled GPS segments, automatically extracting relevant features from 4-channel tensor representations.
\item \textbf{ConvLSTM} \citep{nawaz_convolutional_2020}: A hybrid architecture that uses convolutional layers to extract spatial features from GPS data, followed by LSTM layers to capture temporal patterns, incorporating both location and weather features to enhance mode detection.
\end{itemize}
\bigskip
\noindent
All models were evaluated on the Geolife dataset using identical initial pre-processing and train-test splits. We have also added architectural comparison between our model and Transformer-based Deep-ViT in Appendix~\ref{appendix:parameters}.Table~\ref{tab:model_comparison} presents the test accuracies ranked from highest to lowest.

\begin{table}[t]
\centering
\caption{Test Accuracy Comparison on Geolife (Ordered by Performance)}
\label{tab:model_comparison}
\renewcommand{\arraystretch}{1.1}
\begin{tabular}{lc}
\toprule
\textbf{Model} & \textbf{Test Acc. (\%)} \\
\midrule
\textbf{SpeedTransformer (Ours)} & \textbf{95.97} \\
Deep-ViT \citep{ribeiro_deep_2024} & 92.96 \\
LSTM-based DNN \citep{yu_semi-supervised_2020} & 92.70 \\
LSTM-Attention (Baseline) & \underline{92.40} \\
CE-RCRF \citep{Zeng2023} & 85.23 \\
SECA \citep{dabiri_semi-supervised_2020} & 84.80 \\
ConvLSTM \citep{nawaz_convolutional_2020} & 83.81 \\
\bottomrule
\end{tabular}
\end{table}

\textsc{SpeedTransformer} achieved the highest test accuracy of 95.97\%, outperforming all competing approaches despite using only speed as input. Deep-ViT (92.96\%) and the LSTM-based DNN (92.70\%) achieved strong results but required more complex pre-processing or architectural components. Our LSTM-Attention baseline (92.40\%) demonstrated that while recurrent networks with attention could capture temporal patterns effectively, they still lagged behind the pure attention-based design of \textsc{SpeedTransformer} in predicting complex mobility patterns.

The lower-ranked models illustrated different architectural trade-offs. The CNN-ensemble extracted localized spatial patterns but struggled with sequential dependencies; ConvLSTM improved temporal modeling through its hybrid design but faced generalization challenges; and CE-RCRF treated trajectories as continuous sequences yet was hindered by architectural complexity. We also evaluated a simple rule-based model using the same process and found that it performed substantially worse than all machine learning models (Appendix~\ref{appendix:rule-based}).

\begin{figure}[b]
\centering
\includegraphics[width=1.0\textwidth]{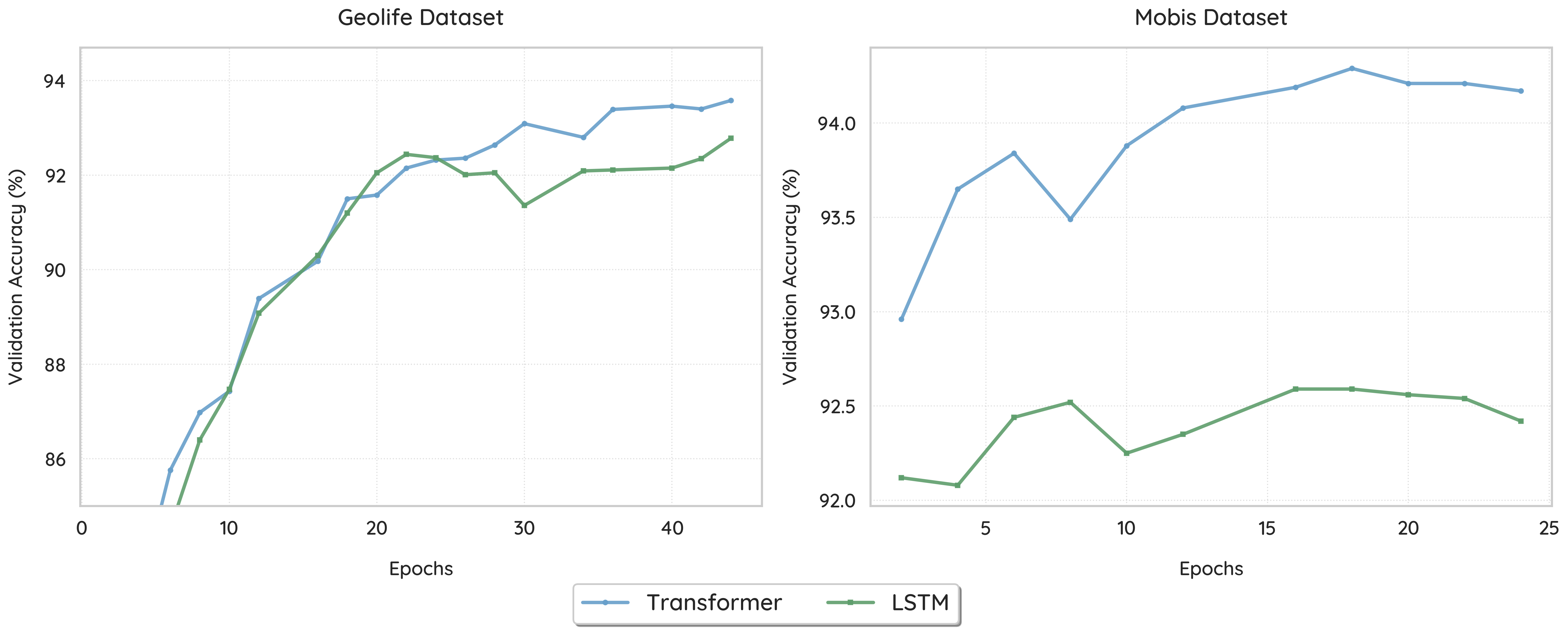}
\caption{Validation accuracies over epochs for Geolife and MOBIS. The SpeedTransformer consistently converges faster and achieves higher overall accuracy than the LSTM-Attention baseline on both datasets.}
\label{fig:validation_accuracies}
\end{figure}

\begin{figure}[b]
\centering
\includegraphics[width=1.0\textwidth]{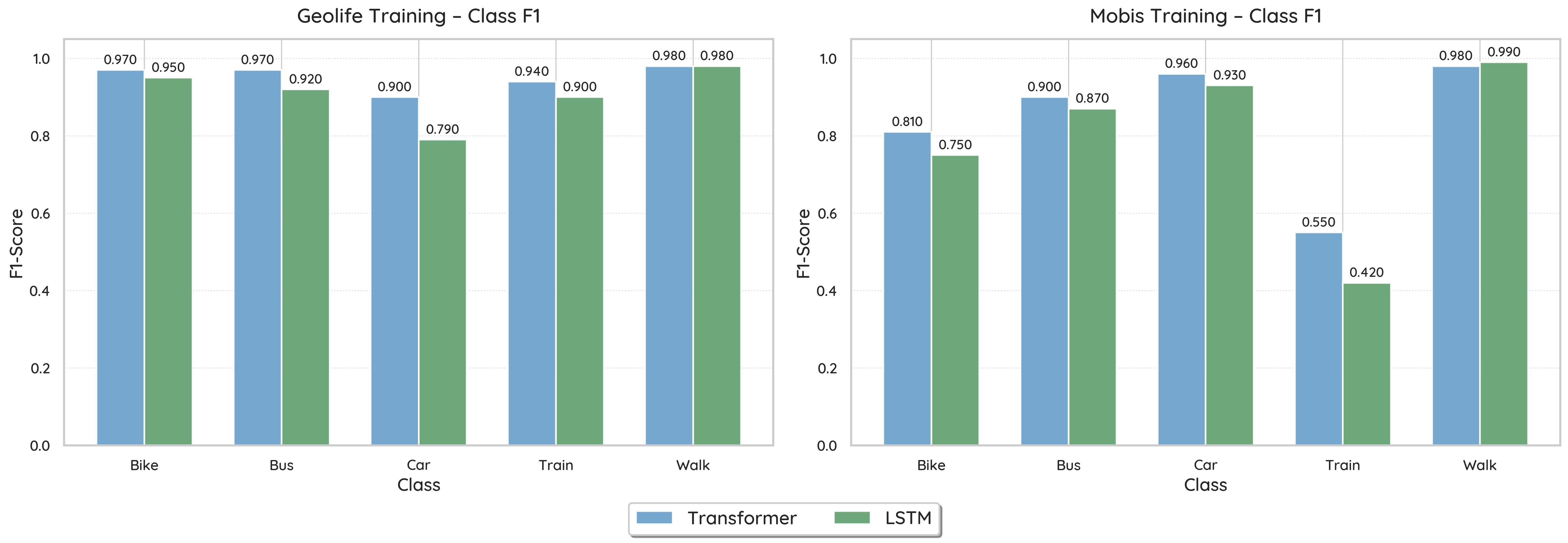}
\caption{Per class F1-Score for Geolife and MOBIS trainings using SpeedTransformer and LSTM. The SpeedTransformer consistently achieves better results than the LSTM-Attention across all classes on both datasets.}
\label{fig:training_f1scores}
\end{figure}

To further assess performance consistency across datasets, we compared \textsc{SpeedTransformer} with our LSTM-Attention baseline on both the Geolife and MOBIS datasets. Figures~\ref{fig:validation_accuracies} and~\ref{fig:training_f1scores} show that \textsc{SpeedTransformer} not only converged faster and achieved higher validation accuracy but also maintained superior F1-scores across all transportation modes. The F1-score in Figure~\ref{fig:training_f1scores}, which examines accuracy by class, provided a more granular and reliable measure of model performance under class imbalance. Moreover, Table~\ref{tab:precision_recall_comparison} presents a detailed comparison of precision and recall between \textsc{SpeedTransformer} and Deep-ViT—the most competitive alternative—demonstrating that \textsc{SpeedTransformer} achieved a more balanced trade-off between precision and recall across datasets, underscoring its robustness and generalizability.


\bigskip
\noindent
\textsc{SpeedTransformer} demonstrated two significant advantages over the LSTM-Attention baseline: faster convergence and higher accuracy across both datasets. As shown in Figure~\ref{fig:training_f1scores}, the model consistently achieved higher validation accuracies throughout training, reaching 94.22\% accuracy on MOBIS compared to 92.33\% for the LSTM-Attention model. The faster convergence was particularly valuable when working with large-scale mobility datasets such as MOBIS, where training efficiency was critical. We also observed that \textsc{SpeedTransformer} achieved superior per-class F1-scores across all transportation modes in both the Geolife and MOBIS datasets. Class-wise error patterns are further summarized by the confusion matrices (Figure~\ref{fig:conf_mats_appendix}) and per-class F1 score (Figure~\ref{fig:transfer}) in Appendix~\ref{appendix:confusion_matrices}. Across different freezing strategies and experimental settings, \textsc{SpeedTransformer} consistently outperforms the LSTM baseline, demonstrating robust performance, as shown in Tables~\ref{table:lstm-mobis}, \ref{table:transformer-mobis}, \ref{table:lstm-geolife}, \ref{table:transformer-geolife}, and \ref{table:LSTM-geolife}. Although the model required GPU resources, it remained relatively efficient, completing both training and inference on a single GPU node (Appendix~\ref{appendix:computational_efficiency}).

\begin{table}[t]
\centering
\caption{Validation Precision and Recall Comparison between Deep-ViT and SpeedTransformer on Geolife}
\label{tab:precision_recall_comparison}
\renewcommand{\arraystretch}{1.1}
\begin{tabular}{lcccc}
\toprule
\multirow{2}{*}{\textbf{Mode}} & \multicolumn{2}{c}{\textbf{Deep-ViT}} & \multicolumn{2}{c}{\textbf{SpeedTransformer}} \\
\cmidrule(lr){2-3} \cmidrule(lr){4-5}
 & \textbf{Precision (\%)} & \textbf{Recall (\%)} & \textbf{Precision (\%)} & \textbf{Recall (\%)} \\
\midrule
Walk  & 95.04 & 96.12 & \underline{96.72} & \underline{99.90} \\
Bike  & 93.79 & 91.40 & \underline{98.54} & \underline{95.45} \\
Bus   & 89.68 & 91.13 & \underline{98.14} & \underline{95.49} \\
Car   & \underline{88.96}& 87.26 & 87.33 & \underline{93.65} \\
Train & 90.00 & 79.41 & \underline{95.66} & \underline{92.93} \\
\midrule
\textbf{Macro Avg.} & 91.89 & 89.86 & \textbf{95.68} & \textbf{95.08} \\
\bottomrule
\end{tabular}
\end{table}

\subsection{Cross-Regional Transferability}

Learning from raw trajectories was feasible for all models, including rule-based approaches. However, deep learning architectures such as Transformers achieved state-of-the-art accuracy through inductive transfer and fine-tuning of pretrained models \citep{howard2018universal}. This transfer-learning approach was particularly advantageous when cross-regional differences in transportation networks and traffic dynamics produced distinct patterns of transportation modes. To evaluate model transferability, we fine-tuned the MOBIS-pretrained models on small subsets of the Geolife dataset to simulate low-shot adaptation. Both \textsc{SpeedTransformer} and LSTM-Attention were fine-tuned on 100 and 200 Geolife trips (approximately 1.1\% and 2.2\% of the dataset, respectively) for 20 epochs and were subsequently evaluated on the remaining samples. Table~\ref{tab:Geolife_small_subsets} reports the overall classification accuracy.

\begin{table}[t]
\centering
\caption{Model Accuracy during Cross-Dataset Fine-Tuning from MOBIS to Small Geolife Subsets}
\label{tab:Geolife_small_subsets}
\begin{tabular}{@{}lcc@{}}
\toprule
\textbf{Model} & \textbf{100 trips} & \textbf{200 trips} \\
\midrule
LSTM-Attention & 75.47\% & 79.15\% \\
SpeedTransformer & 80.53\% & 86.13\% \\
\bottomrule
\end{tabular}
\end{table}
\bigskip
\noindent

Fine-tuning followed a standardized low-shot protocol using MOBIS-pretrained checkpoints. For \textsc{SpeedTransformer}, the encoder and attention layers were frozen, and training was performed with mixed precision for up to 20 epochs. The LSTM-Attention baseline was fine-tuned with smaller batch sizes, a lower learning rate, stronger regularization (dropout = 0.3, weight decay = 5e-3), and gradient clipping at 0.25. Hyperparameters were selected through a grid search to ensure optimal stability under limited supervision (see Appendix~\ref{appendix:gridsearch}). \textsc{SpeedTransformer} achieved 80.53\% accuracy with only 100 trips, surpassing LSTM-Attention by more than five percentage points, and further improved to 86.13\% with 200 trips. These results confirmed its superior capacity to transfer learned mobility representations across regions with minimal labeled data.

\section{Real-World Field Experiment}

Having established \textsc{SpeedTransformer}’s superior accuracy, minimal input requirements, and strong cross-regional transferability, we next evaluated its robustness under real-world conditions, where GPS data were inherently noisy, irregular, and device-dependent. While curated benchmarks provided clean and controlled comparisons, they did not capture the complexities of real-world mobility data, which were subject to signal loss, user heterogeneity, and hardware variability. To address this gap, we conducted a large-scale field experiment to assess \textsc{SpeedTransformer}’s reliability in real-world environments characterized by high uncertainty and unpredictability.

\subsection{Smartphone Application and Data Collection}

\begin{figure}[b]
\centering
\begin{tabular}{@{}c@{\hspace{4pt}}c@{\hspace{4pt}}c@{}}
\includegraphics[width=0.25\textwidth]{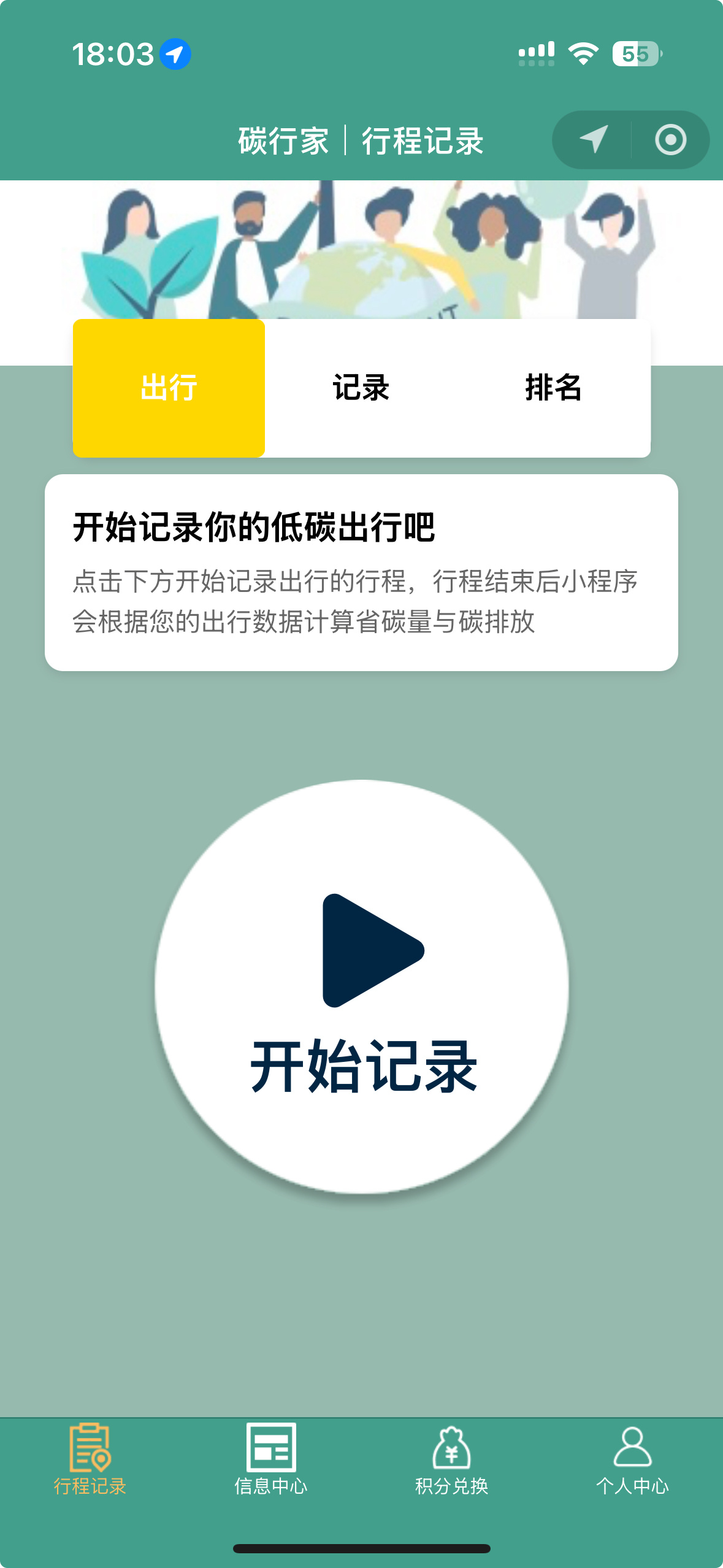} &
\includegraphics[width=0.25\textwidth]{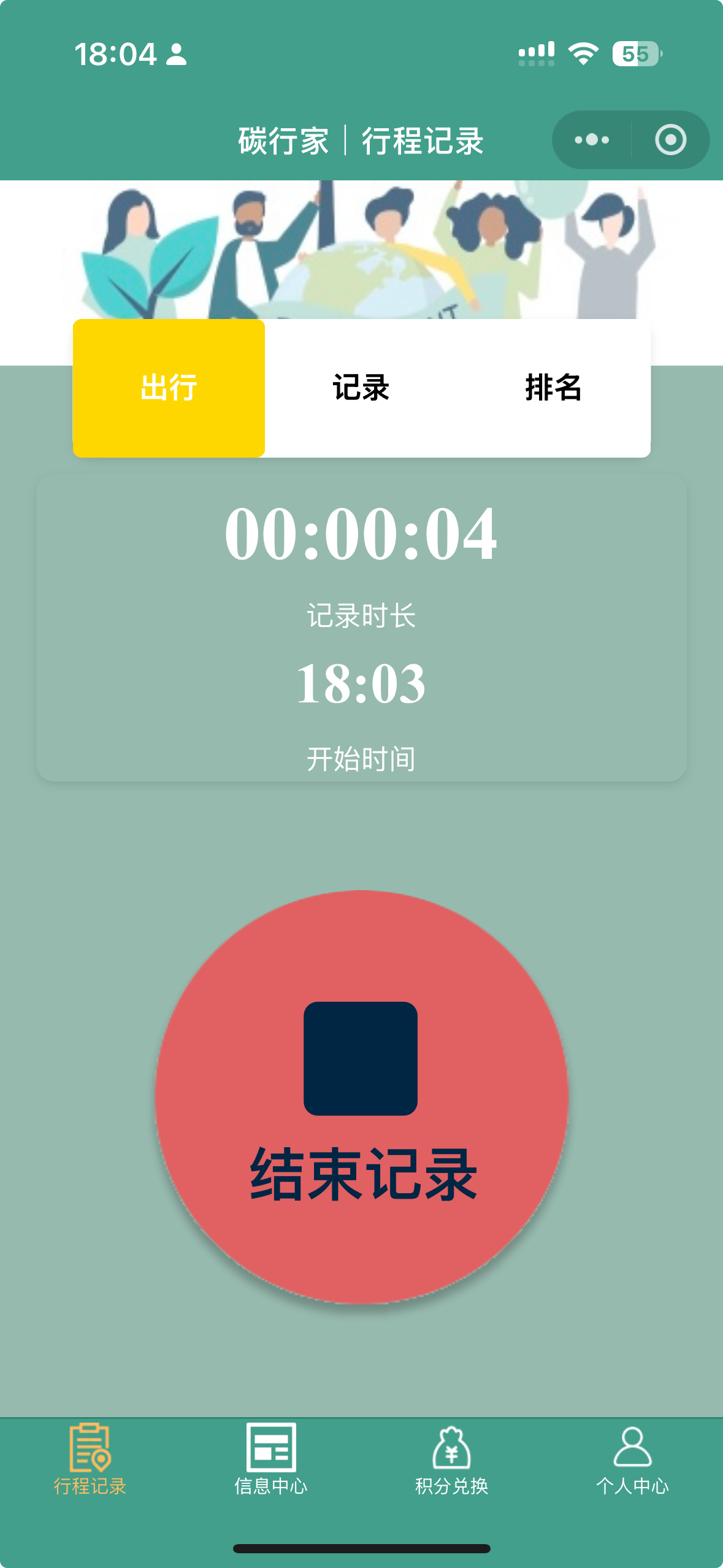} &
\includegraphics[width=0.25\textwidth]{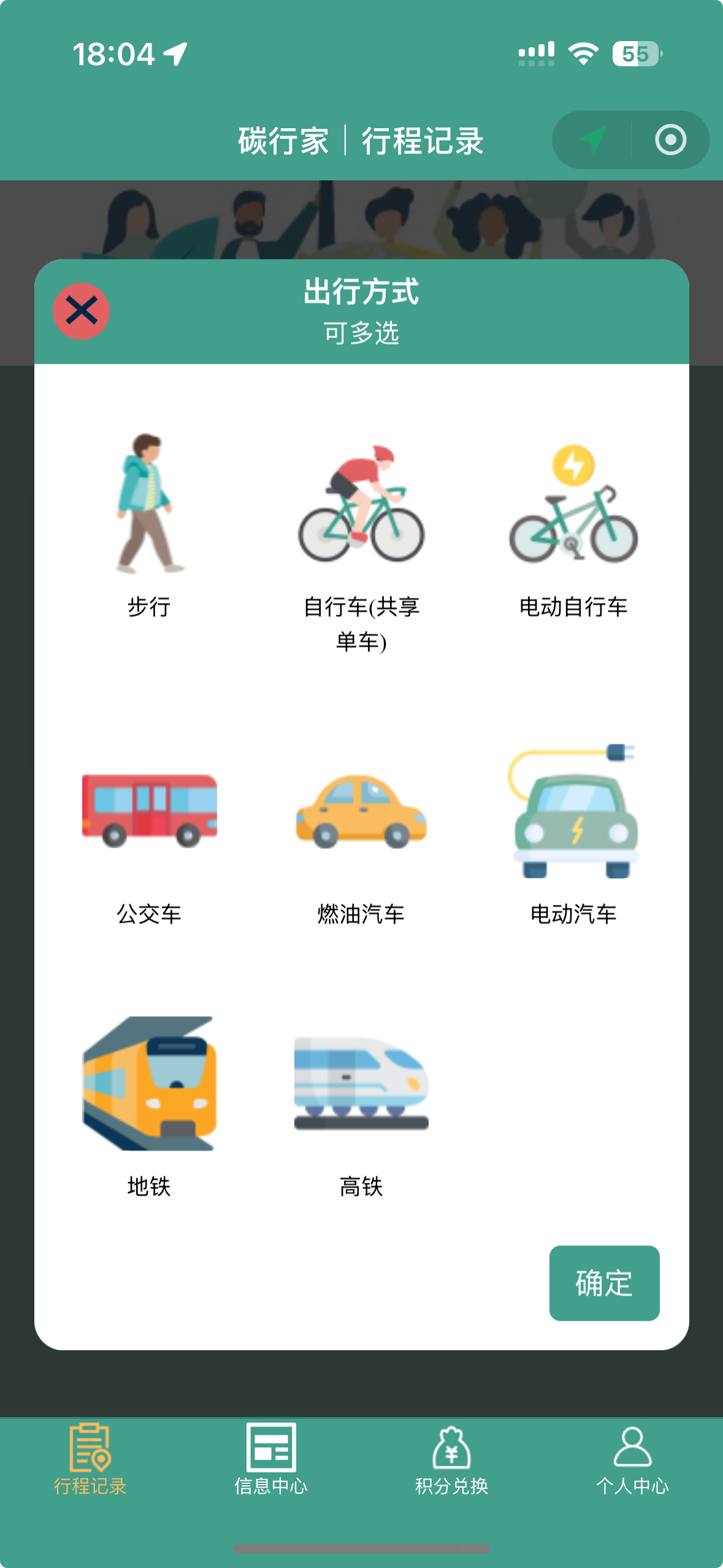} \\
(a) Tracking initiation & (b) Active tracking & (c) Trip completion \\
\end{tabular}
\caption{Trip tracking interface in the CarbonClever application: (a) Trip initiation screen, where users start a new trip recording. (b) Active tracking screen showing real-time trip duration and status. (c) Trip completion and mode confirmation screen, where users select and verify the transportation mode. All interface texts are originally in Chinese.}
\label{fig:user_tracking}
\end{figure}

We developed \textit{CarbonClever}, a WeChat-integrated mini-program designed to estimate individual carbon footprints through continuous mobility tracking.\footnote{A WeChat mini-program is a lightweight application embedded within the WeChat ecosystem. Functionally similar to a simplified smartphone app, the mini-program operates on top of the WeChat super-app, which is ubiquitously used in China. This integration enabled efficient participant recruitment and minimized testing costs across different mobile operating systems and device platforms.} The application provided a streamlined interface for trip initiation, real-time monitoring, and post-trip mode verification (Figure~\ref{fig:user_tracking}). A total of 348 participants from Jiangsu, China, were recruited through an environmental organization to record their daily movements.\footnote{The experiment was approved by the Duke Kunshan University Institutional Review Board (protocol 2022CC073). See Appendix~\ref{irb} for protocol summary.}

\begin{figure}[t]
\centering
\includegraphics[width=1\textwidth]{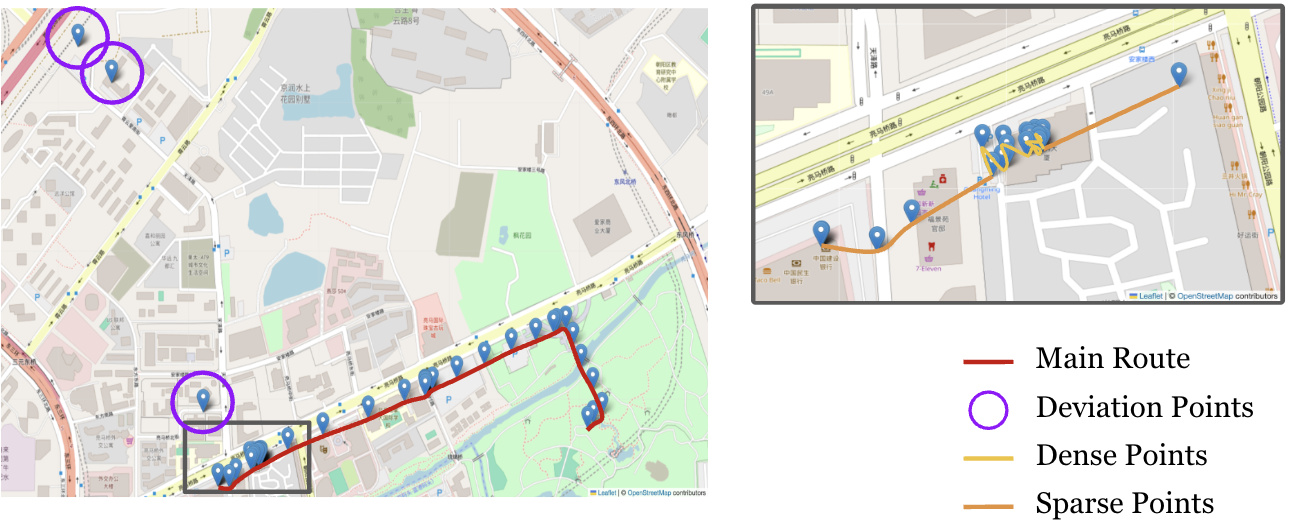}
\caption{Example of real-world GPS trajectories. The red line indicates the main route, with varying point density reflecting heterogeneous sampling frequencies. Purple circles mark signal interference and positioning noise.}
\label{fig:gps_outlier_graph}
\end{figure}

Real-world GPS traces differed substantially from those collected under laboratory conditions. The sampling frequencies varied widely: older iPhone SE models recorded locations every 30--60\,s, whereas newer iPhone 14 Pro devices achieved 5--10\,s intervals under identical conditions (see Appendix~\ref{appendix:temporal} for full temporal sampling comparison). As shown in Figure~\ref{fig:gps_outlier_graph}, trajectories exhibited irregular sampling densities, signal dropouts, and spurious positional jumps caused by multi-path effects or indoor transitions. Unlike benchmark datasets, we intentionally retained these imperfections to evaluate the model’s robustness under realistic deployment conditions.

\begin{table}[ht]
\centering
\caption{Summary of the Mini-Program Dataset}
\label{tab:mini_program_summary}
\begin{tabular}{@{}lrr@{}}
\toprule
\textbf{Mode}    & \textbf{Number of Data Points} & \textbf{Number of Unique Trips} \\ \midrule
Walk   & 28{,}985 & 100 \\
Bus    & 32{,}547 & 259 \\
Car    & 40{,}885 & 205 \\
Bike   & 4{,}901  & 36 \\
Train  & 1{,}505  & 49 \\
\midrule
\textbf{Total} & \textbf{108{,}823} & \textbf{649} \\
\bottomrule
\end{tabular}
\end{table}

\subsection{Field Experiment Evaluation}

We conducted a one-month field experiment from November 18, 2023 to December 23, 2024 that collected 649 verified trips totaling 108{,}823 GPS points from heterogeneous devices across both iOS and Android platforms, ensuring representative variability in sampling rates and noise profiles (see Appendix ~\ref{gps-tracking} for detailed data pre-processing). To assess real-world adaptability, we fine-tuned MOBIS-pretrained SpeedTransformer and LSTM-Attention models using progressively larger subsets of the collected data. Each model was trained for up to 20 epochs with early stopping and identical hyperparameter configurations derived from the grid search (Appendix~\ref{appendix:gridsearch}).  

\begin{table}[t]
\centering
\caption{Fine-Tuning Accuracies of Models with Real-World Data Subsets (5-fold CV; mean $\pm$ std, \% accuracy)}
\label{tab:finetuning_results}
\renewcommand{\arraystretch}{1.1}
\begin{tabular}{rcc}
\toprule
\textbf{Data Subset (\%)} & \textbf{LSTM-Attention Acc. (\%)} & \textbf{SpeedTransformer Acc. (\%)} \\
\midrule
40\% (251 trips) & $88.79 \pm 2.15$ & \underline{$90.27 \pm 2.04$} \\
50\% (314 trips) & $88.82 \pm 1.32$ & \underline{$89.98 \pm 2.55$} \\
60\% (377 trips) & $89.34 \pm 1.86$ & \underline{$90.40 \pm 2.17$} \\
70\% (440 trips) & $90.04 \pm 2.26$ & \underline{$91.72 \pm 3.27$} \\
80\% (502 trips) & $89.30 \pm 2.52$ & \underline{$90.11 \pm 4.54$} \\
\bottomrule
\end{tabular}
\end{table}

SpeedTransformer consistently outperformed the LSTM-Attention baseline across all real-world training subsets (Table~\ref{tab:finetuning_results}). Results are reported using 5-fold cross-validation, with values shown as mean $\pm$ standard deviation of accuracy across folds. SpeedTransformer achieved its best performance at the 70\% subset ($91.72 \pm 3.27$\%), while LSTM-Attention peaked at $90.04 \pm 2.26$\%. Overall, these real-world results reinforce our earlier findings: SpeedTransformer remains more accurate under genuine, noisy, and device-diverse GPS trajectories, helping bridge the gap between research prototypes and operational smart-mobility applications.

We also expanded our modeling experiments to include a per-class analysis, given that our field-experiment data are imbalanced and contain relatively few observations for bikes and trains. This imbalance negatively affected classification accuracy for these modes, resulting in accuracies below 0.8, as shown in Figure~\ref{fig:transfer}. Nevertheless, the per-class accuracies for bikes and trains still exceed 0.5, while the remaining classes maintain very high accuracies above 0.9.

\FloatBarrier
\section{Discussion and Conclusion}

Our research addressed three fundamental challenges in transportation mode detection: (1) the reliance on extensive feature engineering and input pre-processing, (2) limited model generalizability across distinct geographical contexts, and (3) insufficient and unpredictable real-world validation. By achieving state-of-the-art performance in transportation mode detection, our work demonstrated that Transformer-based neural network architectures, when coupled with high-quality training data, could substantially improve mode detection from GPS trajectories. More broadly, this approach had the potential to extract information from a wide range of mobility-based sequence data, and it contributed to a growing body of literature on transportation mode detection and its downstream applications in transportation research, GIS, urban analytics, and climate change science. 

Compared to classical and deep ML methods \citep{ribeiro_deep_2024, Zeng2023, yu_semi-supervised_2020, dabiri_semi-supervised_2020, nawaz_convolutional_2020}, \textsc{SpeedTransformer} outperformed feature-rich models despite using only speed as input. Across different hyperparameter configurations, our model consistently outperformed the LSTM-Attention baseline by 2–3\% (Appendix~\ref{appendix:gridsearch}) and surpassed several other machine learning models by more than 10\% (Table~\ref{tab:model_comparison}). It also outperformed the classical rule-based model, which relied solely on speed, by over 30\% (Appendix~\ref{appendix:rule-based}). This suggests that the Transformer-based model is highly effective at extracting useful information from dense speed sequences that would otherwise underutilized, and it outperforms simpler deterministic model, classical machine-learning models, as well as other deep learning models.

The model’s strong performance was largely attributable to its attention mechanism (Appendix~\ref{appendix:old-transformer}). Nevertheless, our integration of \textsc{SwiGLU} activation, Grouped-Query Attention, and pre-attention Layer Normalization yielded modest yet consistent performance improvements, aligning with the findings of \citet{Shazeer2020GLU} and \citet{touvron2023llama}. Consistent with \citet{yu2023graph}, our results further indicated that deeper neural network architectures were better suited to capturing the semantic and sequential structures embedded within GPS trajectories.

Moreover, this strong performance relied solely on instantaneous speed, without requiring GPS coordinates or additional engineered features. 
This architectural choice offered several advantages: it mitigates the direct exposure of sensitive location data by decoupling motion dynamics from absolute
geographic coordinates\citep{thompson2022twelve, de2013unique}. Specifically, we quantify this privacy benefit by contrasting the information entropy required to resolve a spatial position versus that of a speed observation. Because speed is a kinematic scalar inherently decoupled from raw geographic coordinates, it significantly limits the information available to a data-profiling adversary (for detailed discussion see Appendix \ref{appendix:privacy}. In addition, using only speed as model input simplified preprocessing procedures, thereby enhancing reproducibility across various mobile devices and geographical contexts \citep{bolbol2012inferring}.

Furthermore, \textsc{SpeedTransformer} exhibited strong cross-regional transferability. In the out-of-domain evaluation, it surpassed the LSTM-Attention model by 7\% (Table~\ref{tab:Geolife_small_subsets}), demonstrating robust generalizability across distinct geographical and transportation contexts—from Swiss transportation systems (MOBIS) to Chinese urban environment (Geolife). Travel behaviors varied substantially between these regions, and the two datasets were collected nearly a decade apart, introducing considerable differences in travel patterns across modes such as train, cycling, and automobile use. Remarkably, with only 100 fine-tuning trips, \textsc{SpeedTransformer} achieved satisfactory accuracy (80.53\%) on Geolife after pretraining on MOBIS. These findings suggested that \textsc{SpeedTransformer} captured fundamental patterns of human mobility that persisted across infrastructural settings and regional mobility cultures.

Finally, to bridge the gap between research prototypes and real-world implementation, we validated \textsc{SpeedTransformer} in a large-scale real-world experiment via our self designed \textit{CarbonClever} WeChat Mini-Program on personal smartphones. This mobile application collected GPS trajectories from 348 participants across diverse smartphones and operating systems in Jiangsu. This validation provides a third geographic context, representing mid-tier Asian cities that differ significantly in infrastructural scale and traffic dynamics from the European (MOBIS) and megacity (Geolife) environments. This dataset included unpredictable real world conditions, irregular sampling intervals, signal interferences and device-specific noises. 
\textsc{SpeedTransformer} consistently outperformed the LSTM-Attention baseline across all fine-tuning results, achieving 94.22\% accuracy with 50\% of the real-world training data, compared to 87.57\% for LSTM-Attention. While accuracy decreased slightly due to data irregularities, \textsc{SpeedTransformer} retained high stability across transportation modes, confirming its robustness for practical deployment.

A key limitation of Transformer-based models, including \textsc{SpeedTransformer}, was their sensitivity to training data quality. As shown in Table~\ref{tab:precision_recall_comparison} and Figure~\ref{fig:training_f1scores}, the model performed less effectively for the \textit{Train} class in the MOBIS dataset, where data imbalance and under-representation reduced its ability to generalize. As \citet{van2007experimental} noted, class imbalance posed significant challenges to ML accuracy. Still, by pretraining on the larger, more balanced MOBIS dataset and fine-tuning on the smaller, imbalanced Geolife subsets, \textsc{SpeedTransformer} maintained strong predictive accuracy with minimal performance degradation under distributional shift. The model performed even better using our real-world experimental data, achieving an accuracy of 89.12\% with just 94 training trips, compared to 86.13\% in the Geolife fine-tuning task using 200 training trips (Table~\ref{tab:finetuning_results}). These results underscored the model’s adaptability to data-sparse and heterogeneous mobility environments. These results underscored the model’s adaptability to data-sparse and heterogeneous mobility environments.

In addition, despite the privacy advantages of speed-only inputs, the approach is not entirely immune to sophisticated privacy attacks. Although our model avoids collecting raw coordinates, sequential speed profiles can, in some cases, inadvertently reveal identifiable behavioral patterns \citep{laureshyn2009speed}.\footnote{In our real-world experiment, however, all participants lived in dense urban areas, where individual speed profiles were effectively unrecognizable among hundreds of thousands of co-residents.} If an adversary possesses extensive side-channel information, such as high-resolution traffic flow databases or signal timing schedules, they could theoretically attempt to match a specific speed profile to a known corridor, potentially inferring coarse-grained route information. Future work can investigate the integration of enhanced techniques, such as differential privacy and federated learning, to provide formal guarantees against such attacks.

We also conducted a targeted failure analysis (Appendix~\ref{appendix:failure_analysis}) focusing on the \textit{Train} class, whose accuracy is notably lower than that of other classes. In MOBIS, the train category encompasses multiple train modes—such as underground subway, tram, and metro-style services—and is sometimes intermixed with trajectories that resemble car or bus travel. As a result, the associated speed profiles become difficult to separate. These findings suggest that either more sophisticated models or additional mode-specific training data may be needed to further improve classification performance.

Finally, the inherent opacity of the Transformer architecture poses a limitation regarding its interpretability-specifically the difficulty of attributing learned patterns to interpretable mobility features such as acceleration bursts, stop durations, or route choices. Future research could investigate hybrid architectures that integrate Transformer attention mechanisms with physically interpretable mobility features and extend the framework to incorporate contextual signals—such as weather, traffic density, and land use—for downstream mobility tasks, including origin–destination flow estimation and route-level analysis. In addition, expanding evaluation to regions with different road infrastructures and behavioral norms could further validate model universality.











\smallskip
\noindent
The original datasets of MOBIS and Geolife can be found at their respective project websites: the MOBIS dataset (\href{https://www.research-collection.ethz.ch/handle/20.500.11850/553990}{https://www.research-collection.ethz.ch/handle/20.500.11850/553990}) and the Geolife dataset (\href{https://www.microsoft.com/en-us/research/publication/Geolife-gps-trajectory-dataset-user-guide/}{https://www.microsoft.com/en-us/research/publication/Geolife-gps-trajectory-dataset-user-guide/}), both accessible on May 26, 2025.

\section*{Acknowledgments}
We thank the Institute for Transport Planning and Systems at ETH Zürich for providing the data used in our model training. We are also grateful to Yucen Xiao, Peilin He, Zhixuan Lu, Yili Wen, Shanheng Gu, Ziyue Zhong, and Ni Zheng for their excellent research assistance. This project was supported by the 2024 Kunshan Municipal Key R\&D Program under the Science and Technology Special Project (No.~KS2477). ChatGPT~5 was used solely for grammar checking. All remaining errors are our own.

\section*{Disclosure statement}
No potential conflict of interest was reported by the author(s).

\section*{Data and codes availability statement}
Replication data and code can be found at: \url{https://github.com/othmaneechc/SpeedTransformer}. The repository provides source code for data processing, model training and evaluation.

We have also prepared two Colab notebooks: one with the main results and most appendix results at \url{https://shorturl.at/dzVkb}, and another with the remaining appendix results at \url{https://shorturl.at/XZcB8}.

The original datasets of MOBIS and Geolife can be found at their respective project websites: the MOBIS dataset (\url{https://www.research-collection.ethz.ch/handle/20.500.11850/553990}) and the Geolife dataset (\url{https://www.microsoft.com/en-us/research/publication/Geolife-gps-trajectory-dataset-user-guide/}), both accessed on May 26, 2025.

\section*{Additional information}

\subsection*{Funding}
This project was supported by the 2024 Kunshan Municipal Key R\&D Program under the Science and Technology Special Project (No.~KS2477).

\subsection*{Notes on contributors}

\textbf{Yuandong Zhang}

\noindent\textbf{\textit{Yuandong Zhang}} is a master’s student in Computer Science at the University of California, San Diego, USA. His research interests include deep learning and generative AI. He contributed to conceptualisation, methodology, software, validation, and writing – original draft.

\medskip

\textbf{Othmane Echchabi}

\noindent\textbf{\textit{Othmane Echchabi}} is a master’s student in Computer Science affiliated with Mila – Quebec AI Institute and McGill University, Canada. His research focuses on deep learning and spatial image processing. He contributed to conceptualisation, methodology, software, validation, and writing – original draft.

\medskip

\textbf{Tianshu Feng}

\noindent\textbf{\textit{Tianshu Feng}} is a master’s student in Data Science at the University of Pennsylvania, USA. His interests include data engineering and experimental workflows for machine learning. He contributed to software, data curation, and writing – review \& editing.

\medskip

\textbf{Wenyi Zhang}

\noindent\textbf{\textit{Wenyi Zhang}} is an undergraduate student at Duke Kunshan University, China. Her interests include data processing and applied machine learning. She contributed to software and data curation.

\medskip

\textbf{Hsuai-Kai Liao}

\noindent\textbf{\textit{Hsuai-Kai Liao}} is an undergraduate student at Duke Kunshan University, China. His interests include data science and computational experimentation. He contributed to software and validation.

\medskip

\textbf{Charles Chang}

\noindent\textbf{\textit{Charles Chang}} is an Assistant Professor of Environment and Urban Studies at Duke Kunshan University, China. His research focuses on computational social science, geospatial analysis, and deep learning applications. He contributed to conceptualisation, supervision, funding acquisition, and writing – review \& editing.

\bibliographystyle{unsrtnat}
\bibliography{references}

@preamble{    "\newcommand{\noopsort}[1]{} "
            # "\newcommand{\printfirst}[2]{#1} "
            # "\newcommand{\singleletter}[1]{#1} "
            # "\newcommand{\switchargs}[2]{#2#1} " }

@article{sweeny2002,
author = {Sweeney, Latanya},
title = {k-anonymity: a model for protecting privacy},
year = {2002},
issue_date = {October 2002},
publisher = {World Scientific Publishing Co., Inc.},
address = {USA},
volume = {10},
number = {5},
issn = {0218-4885},
url = {https://doi.org/10.1142/S0218488502001648},
doi = {10.1142/S0218488502001648},
journal = {Int. J. Uncertain. Fuzziness Knowl.-Based Syst.},
month = oct,
pages = {557–570},
numpages = {14}
}

@inproceedings{stenneth_transportation_2011,
	address = {Chicago Illinois},
	title = {Transportation mode detection using mobile phones and {GIS} information},
	isbn = {978-1-4503-1031-4},
	url = {https://dl.acm.org/doi/10.1145/2093973.2093982},
	doi = {10.1145/2093973.2093982},
	language = {en},
	urldate = {2025-02-19},
	booktitle = {Proceedings of the 19th {ACM} {SIGSPATIAL} {International} {Conference} on {Advances} in {Geographic} {Information} {Systems}},
	publisher = {ACM},
	author = {Stenneth, Leon and Wolfson, Ouri and Yu, Philip S. and Xu, Bo},
	month = nov,
	year = {2011},
	pages = {54--63},
}

@article{bolbol_inferring_2012,
	series = {Special {Issue}: {Advances} in {Geocomputation}},
	title = {Inferring hybrid transportation modes from sparse {GPS} data using a moving window {SVM} classification},
	volume = {36},
	issn = {0198-9715},
	url = {https://www.sciencedirect.com/science/article/pii/S0198971512000543},
	doi = {10.1016/j.compenvurbsys.2012.06.001},
	number = {6},
	urldate = {2025-02-19},
	journal = {Computers, Environment and Urban Systems},
	author = {Bolbol, Adel and Cheng, Tao and Tsapakis, Ioannis and Haworth, James},
	month = nov,
	year = {2012},
	pages = {526--537}
}

@inproceedings{zheng_understanding_2008,
	address = {Seoul Korea},
	title = {Understanding mobility based on {GPS} data},
	isbn = {978-1-60558-136-1},
	url = {https://dl.acm.org/doi/10.1145/1409635.1409677},
	doi = {10.1145/1409635.1409677},
	language = {en},
	urldate = {2025-02-19},
	booktitle = {Proceedings of the 10th international conference on {Ubiquitous} computing},
	publisher = {ACM},
	author = {Zheng, Yu and Li, Quannan and Chen, Yukun and Xie, Xing and Ma, Wei-Ying},
	month = sep,
	year = {2008},
	pages = {312--321}
}

@article{jahangiri_applying_2015,
	title = {Applying {Machine} {Learning} {Techniques} to {Transportation} {Mode} {Recognition} {Using} {Mobile} {Phone} {Sensor} {Data}},
	volume = {16},
	issn = {1558-0016},
	url = {https://ieeexplore.ieee.org/document/7063936},
	doi = {10.1109/TITS.2015.2405759},
	number = {5},
	urldate = {2025-02-19},
	journal = {IEEE Transactions on Intelligent Transportation Systems},
	author = {Jahangiri, Arash and Rakha, Hesham A.},
	month = oct,
	year = {2015},
	note = {Conference Name: IEEE Transactions on Intelligent Transportation Systems},
	pages = {2406--2417},
}

@article{zheng_geolife_2010,
	title = {{GeoLife}: {A} collaborative social networking service among user, location and trajectory.},
	volume = {33},
	shorttitle = {{GeoLife}},
	url = {https://citeseerx.ist.psu.edu/document?repid=rep1&type=pdf&doi=24ccdcba118ff9a72de4840efb848c7c852ef247},
	number = {2},
	urldate = {2025-02-19},
	journal = {IEEE Data Eng. Bull.},
	author = {Zheng, Yu and Xie, Xing and Ma, Wei-Ying},
	year = {2010},
	note = {Publisher: Citeseer},
	pages = {32--39}
}

@article{dabiri_inferring_2018,
	title = {Inferring transportation modes from {GPS} trajectories using a convolutional neural network},
	volume = {86},
	issn = {0968-090X},
	url = {https://www.sciencedirect.com/science/article/pii/S0968090X17303509},
	doi = {10.1016/j.trc.2017.11.021},
	urldate = {2025-02-19},
	journal = {Transportation Research Part C: Emerging Technologies},
	author = {Dabiri, Sina and Heaslip, Kevin},
	month = jan,
	year = {2018},
	pages = {360--371}
}

@article{dabiri_semi-supervised_2020,
	title ={Semi-{Supervised} {Deep} {Learning} {Approach} for {Transportation} {Mode} {Identification} {Using} {GPS} {Trajectory} {Data}},
	volume = {32},
	copyright = {https://ieeexplore.ieee.org/Xplorehelp/downloads/license-information/IEEE.html},
	issn = {1041-4347, 1558-2191, 2326-3865},
	url = {https://ieeexplore.ieee.org/document/8632766/},
	doi = {10.1109/TKDE.2019.2896985},
	language = {en},
	number = {5},
	urldate = {2025-02-19},
	journal = {IEEE Transactions on Knowledge and Data Engineering},
	author = {Dabiri, Sina and Lu, Chang-Tien and Healip, Kevin and Reddy, Chandan K.},
	month = may,
	year = {2020},
	pages = {1010--1023}
}

@article{dutta_inferencing_2023,
	title = {Inferencing transportation mode using unsupervised deep learning approach exploiting {GPS} point-level characteristics},
	volume = {53},
	issn = {1573-7497},
	url = {https://doi.org/10.1007/s10489-022-04140-9},
	doi = {10.1007/s10489-022-04140-9},
	language = {en},
	number = {10},
	urldate = {2025-02-19},
	journal = {Applied Intelligence},
	author = {Dutta, Sumanto and Patra, Bidyut Kr.},
	month = may,
	year = {2023},
	pages = {12489--12503}
}

@article{jiang_trajectorynet_2017,
    title = {{TrajectoryNet}: An Embedded {GPS} Trajectory Representation for Point-based Classification Using Recurrent Neural Networks},
    author = {Jiang, Xiang and de Souza, Erico N. and Pesaranghader, Ahmad and Hu, Baifan and Silver, Daniel L. and Matwin, Stan},
    journal = {arXiv preprint arXiv:1705.02636},
    year = {2017},
    doi = {10.48550/arXiv.1705.02636},
    url = {http://arxiv.org/abs/1705.02636}
}

@article{yu_semi-supervised_2020,
	title = {Semi-supervised deep ensemble learning for travel mode identification},
	volume = {112},
	issn = {0968-090X},
	url = {https://www.sciencedirect.com/science/article/pii/S0968090X19309416},
	doi = {10.1016/j.trc.2020.01.003},
	urldate = {2025-02-19},
	journal = {Transportation Research Part C: Emerging Technologies},
	author = {Yu, James J. Q.},
	month = mar,
	year = {2020},
	pages = {120--135},
}

@article{ribeiro_deep_2024,
	title = {A deep learning approach for transportation mode identification using a transformation of {GPS} trajectory data features into an image representation},
	issn = {2364-4168},
	url = {https://doi.org/10.1007/s41060-024-00510-3},
	doi = {10.1007/s41060-024-00510-3},
	language = {en},
	urldate = {2025-02-19},
	journal = {International Journal of Data Science and Analytics},
	author = {Ribeiro, Ricardo and Trifan, Alina and Neves, António J. R.},
	month = feb,
	year = {2024},
}

@inproceedings{liang_trajformer_2022,
    title = {{TrajFormer}: Efficient Trajectory Classification with Transformers},
    author = {Liang, Yuxuan and Ouyang, Kun and Wang, Yiwei and Liu, Xu and Chen, Hongyang and Zhang, Junbo and Zheng, Yu and Zimmermann, Roger},
    booktitle = {Proceedings of the 31st ACM International Conference on Information \& Knowledge Management},
    series = {CIKM '22},
    year = {2022},
    pages = {1229--1237},
    publisher = {Association for Computing Machinery},
    address = {New York, NY, USA},
    doi = {10.1145/3511808.3557481},
    isbn = {9781450392365},
    location = {Atlanta, GA, USA}
}

@inproceedings{hong_how_2022,
	title = {How do you go where? {Improving} next location prediction by learning travel mode information using transformers},
	shorttitle = {How do you go where?},
	url = {http://arxiv.org/abs/2210.04095},
	doi = {10.1145/3557915.3560996},
	urldate = {2025-02-19},
	booktitle = {Proceedings of the 30th {International} {Conference} on {Advances} in {Geographic} {Information} {Systems}},
	author = {Hong, Ye and Martin, Henry and Raubal, Martin},
	month = nov,
	year = {2022},
	note = {arXiv:2210.04095 [cs]},
	pages = {1--10},
}

@article{vaswani_attention_2017,
    title = {Attention Is All You Need},
    author = {Vaswani, Ashish and Shazeer, Noam and Parmar, Niki and Uszkoreit, Jakob and Jones, Llion and Gomez, Aidan N. and Kaiser, Lukasz and Polosukhin, Illia},
    journal = {arXiv preprint arXiv:1706.03762},
    year = {2017},
    doi = {10.48550/arXiv.1706.03762},
    url = {http://arxiv.org/abs/1706.03762}
}

@article{dosovitskiy_image_2021,
    title = {An Image is Worth 16x16 Words: Transformers for Image Recognition at Scale},
    author = {Dosovitskiy, Alexey and Beyer, Lucas and Kolesnikov, Alexander and Weissenborn, Dirk and Zhai, Xiaohua and Unterthiner, Thomas and Dehghani, Mostafa and Minderer, Matthias and Heigold, Georg and Gelly, Sylvain and Uszkoreit, Jakob and Houlsby, Neil},
    journal = {arXiv preprint arXiv:2010.11929},
    year = {2021},
    doi = {10.48550/arXiv.2010.11929},
    url = {http://arxiv.org/abs/2010.11929},
    note = {ICLR 2021}
}

@article{graser_mobilitydl_2024,
    title = {{MobilityDL}: A Review of Deep Learning From Trajectory Data},
    author = {Graser, Anita and Jalali, Anahid and Lampert, Jasmin and Weißenfeld, Axel and Janowicz, Krzysztof},
    journal = {arXiv preprint arXiv:2402.00732},
    year = {2024},
    doi = {10.48550/arXiv.2402.00732},
    url = {http://arxiv.org/abs/2402.00732},
    note = {Submitted to Geoinformatica}
}

@article{nawaz_convolutional_2020,
	title = {Convolutional {LSTM} based transportation mode learning from raw {GPS} trajectories},
	volume = {14},
	copyright = {© 2020 The Institution of Engineering and Technology},
	issn = {1751-9578},
	url = {https://onlinelibrary.wiley.com/doi/abs/10.1049/iet-its.2019.0017},
	doi = {10.1049/iet-its.2019.0017},
	language = {en},
	number = {6},
	urldate = {2025-02-19},
	journal = {IET Intelligent Transport Systems},
	author = {Nawaz, Asif and Zhiqiu, Huang and Senzhang, Wang and Hussain, Yasir and Khan, Izhar and Khan, Zaheer},
	year = {2020},
	note = {\_eprint: https://onlinelibrary.wiley.com/doi/pdf/10.1049/iet-its.2019.0017},
	pages = {570--577}
}

@inproceedings{asci_novel_2019,
    title = {A Novel Input Set for {LSTM}-Based Transport Mode Detection},
    author = {Asci, Guven and Guvensan, M. Amac},
    booktitle = {2019 {IEEE} International Conference on Pervasive Computing and Communications Workshops ({PerCom} Workshops)},
    year = {2019},
    pages = {107--112},
    publisher = {IEEE},
    doi = {10.1109/PERCOMW.2019.8730799},
    url = {https://ieeexplore.ieee.org/abstract/document/8730799}
}

@article{hochreiter_long_1997,
	title = {Long {Short}-{Term} {Memory}},
	volume = {9},
	issn = {0899-7667},
	url = {https://doi.org/10.1162/neco.1997.9.8.1735},
	doi = {10.1162/neco.1997.9.8.1735},
	number = {8},
	urldate = {2025-02-19},
	journal = {Neural Comput.},
	author = {Hochreiter, Sepp and Schmidhuber, Jürgen},
	month = nov,
	year = {1997},
	pages = {1735--1780},
}

@article{yao2020understanding,
  title={Understanding human activity and urban mobility patterns from massive cellphone data: Platform design and applications},
  author={Yao, Zhenxing and Zhong, Yu and Liao, Qiang and Wu, Jie and Liu, Haode and Yang, Fei},
  journal={IEEE Intelligent Transportation Systems Magazine},
  volume={13},
  number={3},
  pages={206--219},
  year={2020},
  publisher={IEEE}
}

@article{gonzalez2008understanding,
  title={Understanding individual human mobility patterns},
  author={Gonzalez, Marta C and Hidalgo, Cesar A and Barabasi, Albert-Laszlo},
  journal={nature},
  volume={453},
  number={7196},
  pages={779--782},
  year={2008},
  publisher={Nature Publishing Group UK London}
}

@article{bonaccorsi2020economic,
  title={Economic and social consequences of human mobility restrictions under COVID-19},
  author={Bonaccorsi, Giovanni and Pierri, Francesco and Cinelli, Matteo and Flori, Andrea and Galeazzi, Alessandro and Porcelli, Francesco and Schmidt, Ana Lucia and Valensise, Carlo Michele and Scala, Antonio and Quattrociocchi, Walter and others},
  journal={Proceedings of the national academy of sciences},
  volume={117},
  number={27},
  pages={15530--15535},
  year={2020},
  publisher={National Acad Sciences}
}

@article{barbosa2021uncovering,
  title={Uncovering the socioeconomic facets of human mobility},
  author={Barbosa, Hugo and Hazarie, Surendra and Dickinson, Brian and Bassolas, Aleix and Frank, Adam and Kautz, Henry and Sadilek, Adam and Ramasco, Jos{\'e} J and Ghoshal, Gourab},
  journal={Scientific reports},
  volume={11},
  number={1},
  pages={8616},
  year={2021},
  publisher={Nature Publishing Group UK London}
}

@article{mcmichael2020human,
  title={Human mobility, climate change, and health: Unpacking the connections},
  author={McMichael, Celia},
  journal={The Lancet Planetary Health},
  volume={4},
  number={6},
  pages={e217--e218},
  year={2020},
  publisher={Elsevier}
}

@article{guo2015mobile,
  title={Mobile crowd sensing and computing: The review of an emerging human-powered sensing paradigm},
  author={Guo, Bin and Wang, Zhu and Yu, Zhiwen and Wang, Yu and Yen, Neil Y and Huang, Runhe and Zhou, Xingshe},
  journal={ACM computing surveys (CSUR)},
  volume={48},
  number={1},
  pages={1--31},
  year={2015},
  publisher={ACM New York, NY, USA}
}

@article{molloy_mobis_2023,
	title = {The {MOBIS} dataset: a large {GPS} dataset of mobility behaviour in {Switzerland}},
	volume = {50},
	issn = {1572-9435},
	url = {https://doi.org/10.1007/s11116-022-10299-4},
	doi = {10.1007/s11116-022-10299-4},
	number = {5},
	journal = {Transportation},
	author = {Molloy, Joseph and Castro, Alberto and Götschi, Thomas and Schoeman, Beaumont and Tchervenkov, Christopher and Tomic, Uros and Hintermann, Beat and Axhausen, Kay W.},
	month = oct,
	year = {2023},
	pages = {1983--2007},
}

@article{Barbosa2015,
  author = {Barbosa, H. and de Lima-Neto, F. B. and Evsukoff, A. and others},
  title = {The effect of recency to human mobility},
  journal = {EPJ Data Science},
  year = {2015},
  volume = {4},
  pages = {21},
  doi = {10.1140/epjds/s13688-015-0059-8},
  url = {https://doi.org/10.1140/epjds/s13688-015-0059-8},
  received = {2015-07-24},
  accepted = {2015-11-11},
  published = {2015-12-17}
}

@inproceedings{Preotiuc-Pietro2013,
  author = {Preotiuc-Pietro, Daniel and Cohn, Trevor},
  title = {Mining User Behaviours: A Study of Check-in Patterns in Location Based Social Networks},
  booktitle = {Proceedings of the 3rd Annual ACM Web Science Conference (WebSci 2013)},
  year = {2013},
  doi = {10.1145/2464464.2464479}
}

@article{Gordon2013AutomatedIO,
  title={Automated Inference of Linked Transit Journeys in London Using Fare-Transaction and Vehicle Location Data},
  author={Jason B. Gordon and Harilaos N. Koutsopoulos and Nigel H. M. Wilson and John P. Attanucci},
  journal={Transportation Research Record},
  year={2013},
  volume={2343},
  pages={17 - 24},
  url={https://api.semanticscholar.org/CorpusID:109903480}
}

@article{Rzeszewski2018,
  author = {Rzeszewski, M. and Luczys, P.},
  title = {Care, Indifference and Anxiety—Attitudes toward Location Data in Everyday Life},
  journal = {ISPRS International Journal of Geo-Information},
  year = {2018},
  volume = {7},
  number = {10},
  pages = {383},
  doi = {10.3390/ijgi7100383},
  url = {https://doi.org/10.3390/ijgi7100383}
}

@article{Zhao2016,
  author = {Zhao, Ziliang and Shaw, Shih-Lung and Xu, Yang and Lu, Feng and Chen, Jie and Yin, Ling},
  title = {Understanding the bias of call detail records in human mobility research},
  journal = {International Journal of Geographical Information Science},
  year = {2016},
  volume = {30},
  number = {9},
  pages = {1738--1762},
  doi = {10.1080/13658816.2015.1137298},
  url = {https://doi.org/10.1080/13658816.2015.1137298}
}

@article{Pappalardo2023,
  author = {Pappalardo, Luca and Manley, Ed and Sekara, Vedran and Alessandretti, Laura},
  title = {Future directions in human mobility science},
  journal = {Nature Computational Science},
  year = {2023},
  volume = {3},
  number = {7},
  pages = {588--600},
  doi = {10.1038/s43588-023-00469-4},
  url = {https://doi.org/10.1038/s43588-023-00469-4},
  issn = {2662-8457},
  date = {2023/07/01}
}

@techreport{Manzoni2010,
  author = {Manzoni, V. and Manilo, D. and Kloeckl, K. and Ratti, C.},
  title = {Transportation Mode Identification and Real-Time CO2 Emission Estimation Using Smartphones: How CO2GO Works - Technical Report},
  institution = {SENSEable City Lab, Massachusetts Institute of Technology and Politecnico di Milano},
  year = {2010},
  url = {http://senseable.mit.edu/co2go/images/co2go-technical-report.pdf},
}

@article{Xiao2025,
  author = {Xiao, XueFei and Li, ChunHua and Wang, XingJie and Zeng, AnPing},
  title = {Personalized tourism recommendation model based on temporal multilayer sequential neural network},
  journal = {Scientific Reports},
  year = {2025},
  volume = {15},
  number = {1},
  pages = {382},
  doi = {10.1038/s41598-024-84581-z},
  url = {https://doi.org/10.1038/s41598-024-84581-z}
}

@article{Prelipcean2016,
  author = {Prelipcean, Adrian C. and Gidófalvi, Gyözö and Susilo, Yusak O.},
  title = {Transportation Mode Detection – an in-Depth Review of Applicability and Reliability},
  journal = {Transport Reviews},
  year = {2016},
  volume = {37},
  number = {4},
  pages = {442--464},
  doi = {10.1080/01441647.2016.1246489}
}

@article{Aleta2022,
  author = {Aleta, A. and Martín-Corral, D. and Bakker, M. A. and Pastore Y Piontti, A. and Ajelli, M. and Litvinova, M. and Chinazzi, M. and Dean, N. E. and Halloran, M. E. and Longini, I. M. Jr and Pentland, A. and Vespignani, A. and Moreno, Y. and Moro, E.},
  title = {Quantifying the importance and location of SARS-CoV-2 transmission events in large metropolitan areas},
  journal = {Proceedings of the National Academy of Sciences of the United States of America},
  year = {2022},
  volume = {119},
  number = {26},
  pages = {e2112182119},
  doi = {10.1073/pnas.2112182119},
  pmid = {35696558},
  pmcid = {PMC9245708},
  eprint = {https://www.pnas.org/content/119/26/e2112182119.full.pdf},
  url = {https://doi.org/10.1073/pnas.2112182119}
}

@inproceedings{BERT,
    title = "{BERT}: Pre-training of Deep Bidirectional Transformers for Language Understanding",
    author = "Devlin, Jacob  and
      Chang, Ming-Wei  and
      Lee, Kenton  and
      Toutanova, Kristina",
    editor = "Burstein, Jill  and
      Doran, Christy  and
      Solorio, Thamar",
    booktitle = "Proceedings of the 2019 Conference of the North {A}merican Chapter of the Association for Computational Linguistics: Human Language Technologies, Volume 1 (Long and Short Papers)",
    month = jun,
    year = "2019",
    address = "Minneapolis, Minnesota",
    publisher = "Association for Computational Linguistics",
    url = "https://aclanthology.org/N19-1423/",
    doi = "10.18653/v1/N19-1423",
    pages = "4171--4186"
}

@inproceedings{XueLLM,
author = {Xue, Hao and Voutharoja, Bhanu Prakash and Salim, Flora D.},
title = {Leveraging language foundation models for human mobility forecasting},
year = {2022},
isbn = {9781450395298},
publisher = {Association for Computing Machinery},
address = {New York, NY, USA},
url = {https://doi.org/10.1145/3557915.3561026},
doi = {10.1145/3557915.3561026},
booktitle = {Proceedings of the 30th International Conference on Advances in Geographic Information Systems},
articleno = {90},
numpages = {9},
location = {Seattle, Washington},
series = {SIGSPATIAL '22}
}

@article{GPT2,
  title={Language Models are Unsupervised Multitask Learners},
  author={Radford, Alec and Wu, Jeff and Child, Rewon and Luan, David and Amodei, Dario and Sutskever, Ilya},
  year={2019}
}

@article{Zeng2023,
  author = {Jiaqi Zeng and Yi Yu and Yong Chen and Di Yang and Lei Zhang and Dianhai Wang},
  title = {Trajectory-as-a-Sequence: A novel travel mode identification framework},
  journal = {Transportation Research Part C: Emerging Technologies},
  volume = {146},
  year = {2023},
  pages = {103957},
  issn = {0968-090X},
  doi = {10.1016/j.trc.2022.103957},
  url = {https://www.sciencedirect.com/science/article/pii/S0968090X22003709}
}

@article{Zandbergen2009,
  author = {Zandbergen, Paul A.},
  title = {Accuracy of iPhone Locations: A Comparison of Assisted GPS, WiFi and Cellular Positioning},
  journal = {Transactions in GIS},
  volume = {13},
  pages = {5-25},
  year = {2009},
  publisher = {Wiley},
  doi = {10.1111/j.1467-9671.2009.01152.x},
  url = {https://doi.org/10.1111/j.1467-9671.2009.01152.x}
}

@inproceedings{Cui2001,
  author = {Cui, Y. J. and Ge, S. S.},
  title = {Autonomous vehicle positioning with GPS in urban canyon environments},
  booktitle = {Proceedings 2001 ICRA. IEEE International Conference on Robotics and Automation (Cat. No.01CH37164)},
  location = {Seoul, Korea (South)},
  year = {2001},
  volume = {2},
  pages = {1105-1110},
  doi = {10.1109/ROBOT.2001.932759}
}

@article{lu2012predictability,
  title={Predictability of population displacement after the 2010 Haiti earthquake},
  author={Lu, Xin and Bengtsson, Linus and Holme, Petter},
  journal={Proceedings of the National Academy of Sciences},
  volume={109},
  number={29},
  pages={11576--11581},
  year={2012},
  publisher={National Academy of Sciences}
}

@article{goodchild2013quality,
  title={The quality of big (geo) data},
  author={Goodchild, Michael F},
  journal={Dialogues in Human Geography},
  volume={3},
  number={3},
  pages={280--284},
  year={2013},
  publisher={SAGE Publications Sage UK: London, England}
}

@article{heimgartner2024modal,
  title={Modal splits before, during, and after the pandemic in Switzerland},
  author={Heimgartner, Daniel and Axhausen, Kay W},
  journal={Transportation research record},
  volume={2678},
  number={7},
  pages={1084--1099},
  year={2024},
  publisher={SAGE Publications Sage CA: Los Angeles, CA}
}

@article{zook2015geographies,
  author = {Zook, Matthew and Kraak, Menno-Jan and Ahas, Rein},
  title = {Geographies of mobility: applications of location-based data},
  journal = {International Journal of Geographical Information Science},
  volume = {29},
  number = {11},
  pages = {1935--1940},
  year = {2015},
  publisher = {Taylor \& Francis},
  doi = {10.1080/13658816.2015.1061667},
  url = {https://doi.org/10.1080/13658816.2015.1061667}
}

@article{tao2018analytics,
  title={Analytics of movement through checkpoints},
  author={Tao, Yaguang and Both, Alan and Duckham, Matt},
  journal={International Journal of Geographical Information Science},
  volume={32},
  number={7},
  pages={1282--1303},
  year={2018},
  publisher={Taylor \& Francis}
}

@article{shaw2016human,
  title={Human dynamics in the mobile and big data era},
  author={Shaw, Shih-Lung and Tsou, Ming-Hsiang and Ye, Xinyue},
  journal={International Journal of Geographical Information Science},
  volume={30},
  number={9},
  pages={1687--1693},
  year={2016},
  publisher={Taylor \& Francis}
}

@article{wang2015carbon,
  title={Carbon emission from urban passenger transportation in Beijing},
  author={Wang, Zijia and Chen, Feng and Fujiyama, Taku},
  journal={Transportation Research Part D: Transport and Environment},
  volume={41},
  pages={217--227},
  year={2015},
  publisher={Elsevier}
}

@article{schuessler2009processing,
  title={Processing raw data from global positioning systems without additional information},
  author={Schuessler, Nadine and Axhausen, Kay W},
  journal={Transportation Research Record},
  volume={2105},
  number={1},
  pages={28--36},
  year={2009},
  publisher={SAGE Publications Sage CA: Los Angeles, CA}
}

@article{cheng2016long,
  title={Long short-term memory-networks for machine reading},
  author={Cheng, Jianpeng and Dong, Li and Lapata, Mirella},
  journal={arXiv preprint arXiv:1601.06733},
  year={2016}
}

@article{Su2021RoPE,
  title   = {RoFormer: Enhanced Transformer with Rotary Position Embedding},
  author  = {Su, Jianlin and Lu, Yu and Pan, Shengfeng and Murtadha, Ahmed and Wen, Bo and Liu, Yunfeng},
  journal = {arXiv preprint arXiv:2104.09864},
  year    = {2021},
  doi     = {10.48550/arXiv.2104.09864},
  url     = {https://arxiv.org/abs/2104.09864}
}

@article{Shazeer2020GLU,
  title   = {GLU Variants Improve Transformer},
  author  = {Shazeer, Noam},
  journal = {arXiv preprint arXiv:2002.05202},
  year    = {2020},
  doi     = {10.48550/arXiv.2002.05202},
  url     = {https://arxiv.org/abs/2002.05202}
}

@article{Ainslie2023GQA,
  title   = {GQA: Training Generalized Multi-Query Transformer Models from Multi-Head Checkpoints},
  author  = {Ainslie, Joshua and Lee-Thorp, James and de Jong, Michiel and Zemlyanskiy, Yury and Lebr{\'o}n, Federico and Sanghai, Sumit},
  journal = {arXiv preprint arXiv:2305.13245},
  year    = {2023},
  doi     = {10.48550/arXiv.2305.13245},
  url     = {https://arxiv.org/abs/2305.13245}
}

@inproceedings{xu2017trajectory,
  title={Trajectory recovery from ash: User privacy is not preserved in aggregated mobility data},
  author={Xu, Fengli and Tu, Zhen and Li, Yong and Zhang, Pengyu and Fu, Xiaoming and Jin, Depeng},
  booktitle={Proceedings of the 26th international conference on world wide web},
  pages={1241--1250},
  year={2017},
  doi={https://doi.org/10.1145/3038912.3052620}
}

@article{girod2013influence,
  title={Influence of travel behavior on global CO2 emissions},
  author={Girod, Bastien and van Vuuren, Detlef P and de Vries, Bert},
  journal={Transportation Research Part A: Policy and Practice},
  volume={50},
  pages={183--197},
  year={2013},
  publisher={Elsevier}
}

@article{tajalli2017relationships,
  title={On the relationships between commuting mode choice and public health},
  author={Tajalli, Mehrdad and Hajbabaie, Ali},
  journal={Journal of Transport \& Health},
  volume={4},
  pages={267--277},
  year={2017},
  publisher={Elsevier}
}

@article{de2013unique,
  title={Unique in the crowd: The privacy bounds of human mobility},
  author={De Montjoye, Yves-Alexandre and Hidalgo, C{\'e}sar A and Verleysen, Michel and Blondel, Vincent D},
  journal={Scientific reports},
  volume={3},
  number={1},
  pages={1376},
  year={2013},
  publisher={Nature Publishing Group UK London}
}

@inproceedings{michael2006emerging,
  title={The emerging ethics of humancentric GPS tracking and monitoring},
  author={Michael, Katina and McNamee, Andrew and Michael, Michael G},
  booktitle={2006 International Conference on Mobile Business},
  pages={34--34},
  year={2006},
  organization={IEEE}
}

@inproceedings{klasnja2009exploring,
  title={Exploring privacy concerns about personal sensing},
  author={Klasnja, Predrag and Consolvo, Sunny and Choudhury, Tanzeem and Beckwith, Richard and Hightower, Jeffrey},
  booktitle={International Conference on Pervasive Computing},
  pages={176--183},
  year={2009},
  organization={Springer}
}

@inproceedings{minch2004privacy,
  title={Privacy issues in location-aware mobile devices},
  author={Minch, Robert P},
  booktitle={37th Annual Hawaii International Conference on System Sciences, 2004. Proceedings of the},
  pages={10--pp},
  year={2004},
  organization={IEEE}
}

@article{ng2002price,
  title={The price of convenience: Privacy and mobile commerce},
  author={Ng-Kruelle, Grace and Swatman, Paul A and Rebne, Douglas S and Hampe, J Felix},
  journal={Quarterly journal of electronic commerce},
  volume={3},
  pages={273--286},
  year={2002},
  publisher={INFORMATION AGE PUBLISHING}
}

@article{krumm2009survey,
  title={A survey of computational location privacy},
  author={Krumm, John},
  journal={Personal and Ubiquitous Computing},
  volume={13},
  number={6},
  pages={391--399},
  year={2009},
  publisher={Springer}
}

@article{shin2012privacy,
  title={Privacy protection for users of location-based services},
  author={Shin, Kang G and Ju, Xiaoen and Chen, Zhigang and Hu, Xin},
  journal={IEEE Wireless Communications},
  volume={19},
  number={1},
  pages={30--39},
  year={2012},
  publisher={IEEE}
}

@incollection{thompson2022twelve,
  title={Twelve million phones, one dataset, zero privacy},
  author={Thompson, Stuart A and Warzel, Charlie},
  booktitle={Ethics of data and analytics},
  pages={161--169},
  year={2022},
  publisher={Auerbach Publications}
}

@article{jiang2021location,
  title={Location privacy-preserving mechanisms in location-based services: A comprehensive survey},
  author={Jiang, Hongbo and Li, Jie and Zhao, Ping and Zeng, Fanzi and Xiao, Zhu and Iyengar, Arun},
  journal={ACM Computing Surveys (CSUR)},
  volume={54},
  number={1},
  pages={1--36},
  year={2021},
  publisher={ACM New York, NY, USA}
}

@article{touvron2023llama,
  title={Llama: Open and efficient foundation language models},
  author={Touvron, Hugo and Lavril, Thibaut and Izacard, Gautier and Martinet, Xavier and Lachaux, Marie-Anne and Lacroix, Timoth{\'e}e and Rozi{\`e}re, Baptiste and Goyal, Naman and Hambro, Eric and Azhar, Faisal and others},
  journal={arXiv preprint arXiv:2302.13971},
  year={2023}
}

@article{howard2018universal,
  title={Universal language model fine-tuning for text classification},
  author={Howard, Jeremy and Ruder, Sebastian},
  journal={arXiv preprint arXiv:1801.06146},
  year={2018}
}

@article{drosouli2023tmd,
  title={TMD-BERT: a transformer-based model for transportation mode detection},
  author={Drosouli, Ifigenia and Voulodimos, Athanasios and Mastorocostas, Paris and Miaoulis, Georgios and Ghazanfarpour, Djamchid},
  journal={Electronics},
  volume={12},
  number={3},
  pages={581},
  year={2023},
  publisher={MDPI}
}

@article{huang2019transport,
  title={Transport mode detection based on mobile phone network data: A systematic review},
  author={Huang, Haosheng and Cheng, Yi and Weibel, Robert},
  journal={Transportation Research Part C: Emerging Technologies},
  volume={101},
  pages={297--312},
  year={2019},
  publisher={Elsevier}
}

@article{bolbol2012inferring,
  title={Inferring hybrid transportation modes from sparse GPS data using a moving window SVM classification},
  author={Bolbol, Adel and Cheng, Tao and Tsapakis, Ioannis and Haworth, James},
  journal={Computers, Environment and Urban Systems},
  volume={36},
  number={6},
  pages={526--537},
  year={2012},
  publisher={Elsevier}
}

@inproceedings{van2007experimental,
  title={Experimental perspectives on learning from imbalanced data},
  author={Van Hulse, Jason and Khoshgoftaar, Taghi M and Napolitano, Amri},
  booktitle={Proceedings of the 24th international conference on Machine learning},
  pages={935--942},
  year={2007}
}

@article{yu2023graph,
  title={Graph based embedding learning of trajectory data for transportation mode recognition by fusing sequence and dependency relations},
  author={Yu, Wenhao and Wang, Guanwen},
  journal={International Journal of Geographical Information Science},
  volume={37},
  number={12},
  pages={2514--2537},
  year={2023},
  publisher={Taylor \& Francis}
}

@article{laureshyn2009speed,
  title={From speed profile data to analysis of behaviour: classification by pattern recognition techniques},
  author={Laureshyn, Aliaksei and {\AA}str{\"o}m, Kalle and Brundell-Freij, Karin},
  journal={IATSS research},
  volume={33},
  number={2},
  pages={88--98},
  year={2009},
  publisher={Elsevier}
}

\clearpage
\appendix
\renewcommand{\thetable}{\thesection\arabic{table}}
\renewcommand{\thefigure}{\thesection\arabic{figure}}

\setcounter{table}{0}
\setcounter{figure}{0}

\bigskip
\section{Data Pre-Processing}
\label{pre-processing}
Our pre-processing pipeline transformed raw GPS trajectories from both the Geolife and MOBIS datasets into standardized, analysis-ready data suitable for model training. The process involved several key steps:

\textbf{Data Consolidation}: For Geolife, we first converted the original PLT files into CSV format, while MOBIS data was already structured appropriately. Tables \ref{tab:mobis_summary} and \ref{tab:Geolife_summary} show the initial distribution of transportation modes in the raw datasets.

\textbf{Distance Calculation}: We computed distances between consecutive GPS points using the Haversine formula, which accounts for Earth's curvature:
\begin{equation}
d = 2r \cdot \arcsin\left(\sqrt{\sin^2\left(\frac{\phi_2 - \phi_1}{2}\right) + \cos(\phi_1) \cos(\phi_2) \sin^2\left(\frac{\lambda_2 - \lambda_1}{2}\right)}\right)
\end{equation}
where $\phi$ represents latitude, $\lambda$ represents longitude (both in radians), and $r$ is Earth's radius (6,371,000 meters).

\textbf{Speed Derivation}: We computed speeds by dividing the distance by the time difference between consecutive points, converting to km/h:
\begin{equation}
v = \frac{d}{\Delta t} \cdot 3.6
\end{equation}

\textbf{Label Standardization}: We harmonized transportation mode labels across datasets, consolidating similar modes (e.g., merging 'taxi' with 'car', and 'subway' with 'train') to create consistent categories across datasets.

\textbf{Mode-Specific Filtering}: We applied speed thresholds for each transportation mode (Table \ref{tab:speed_thresholds}) to remove physically implausible values caused by GPS errors or data anomalies.

\textbf{Trip Quality Control}: We removed trips with fewer than three GPS points, as these would depart from real world scenarios and they provide insufficient sequential information for the model to learn meaningful patterns.

This pre-processing reduced the original datasets to five standardized transportation modes (Bike, Bus, Car, Train, and Walk) suitable for cross-dataset modeling. The final datasets used for model training and evaluation contained 144.7 million data points across 1.53 million trips from MOBIS and 4.44 million data points across 9,427 trips from Geolife, as detailed in Table \ref{tab:dataset_comparison}.

\begin{table}[ht]
\centering
\caption{Speed Thresholds for Each Mode}
\label{tab:speed_thresholds}
\begin{tabular}{@{}lrr@{}}
\toprule
\textbf{Mode} & \textbf{Min Speed (km/h)} & \textbf{Max Speed (km/h)} \\ \midrule
Bike & 0.5 & 50 \\
Bus & 1.0 & 120 \\
Car & 3.0 & 180 \\
Train & 3.0 & 350 \\
Walk & 0.1 & 15 \\
\bottomrule
\end{tabular}
\end{table}
\begin{table}[ht]
\centering
\caption{Summary of the MOBIS Dataset}
\label{tab:mobis_summary}
\begin{tabular}{@{}lrr@{}}
\toprule
\textbf{Mode} & \textbf{Number of Data Points} & \textbf{Number of Unique Trips} \\ \midrule
Aerialway & 109 & 7 \\
Airplane & 718,156 & 3,185 \\
Bicycle & 6,568,587 & 41,166 \\
Bus & 7,450,378 & 77,361 \\
Car & 160,348,026 & 553,259 \\
Ferry & 228 & 12 \\
LightRail & 5,824,637 & 39,657 \\
RegionalTrain & 3,674,625 & 22,413 \\
Subway & 425,223 & 6,917 \\
Train & 10,166,760 & 34,064 \\
Tram & 3,591,472 & 32,428 \\
Walk & 56,539,324 & 767,923 \\ \midrule
\textbf{Total} & \textbf{255,307,525} & \textbf{1,578,392} \\ \bottomrule
\end{tabular}
\end{table}
\begin{table}[ht]
\centering
\caption{Summary of the Labeled Geolife Dataset}
\label{tab:Geolife_summary}
\begin{tabular}{@{}lrr@{}}
\toprule
\textbf{Mode} & \textbf{Number of Data Points} & \textbf{Number of Unique Trips} \\ \midrule
Airplane & 9,196 & 14 \\
Bike & 951,350 & 1,555 \\
Boat & 3,565 & 7 \\
Bus & 1,271,062 & 1,851 \\
Car & 512,939 & 782 \\
Motorcycle & 336 & 2 \\
Run & 1,975 & 4 \\
Subway & 309,699 & 613 \\
Taxi & 241,404 & 513 \\
Train & 556,397 & 177 \\
Walk & 1,582,693 & 3,991 \\
\midrule
\textbf{Total} & \textbf{5,440,616} & \textbf{9,509} \\
\bottomrule
\end{tabular}
\end{table}

\section{GPS Tracking on Smartphone Application}
\label{gps-tracking}

Our smartphone application, \textit{CarbonClever} WeChat Mini-Program, implements real-time GPS tracking to collect high-quality mobility data while minimizing battery consumption. When a participant initiates tracking, the application activates the device’s location services using a dynamic sampling strategy designed to balance data density and energy efficiency.

The location tracking module periodically queries the device’s location sensors at intervals ranging from 10 to 30 seconds, depending on device capability and movement speed. Although the application can technically record location data at one-second intervals, a sampling rate of 10–30 seconds effectively preserves the accuracy of speed estimation while preventing excessive battery drain on users’ smartphones.

Each recorded sample captures the user’s current geographic coordinates $(\phi, \lambda)$ and timestamp $(t)$. As new coordinates are received, the application computes the great-circle distance between consecutive points using the Haversine formula~\eqref{eq:haversine}:

\begin{align}
d &= 2r \arcsin\left(\sqrt{\sin^2\left(\frac{\phi_2 - \phi_1}{2}\right) + \cos(\phi_1)\cos(\phi_2)\sin^2\left(\frac{\lambda_2 - \lambda_1}{2}\right)}\right),
\label{eq:haversine}
\end{align}

where $d$ denotes the great-circle distance in meters, $r$ is Earth’s radius ($6{,}371{,}000$ meters), and $\phi_1$, $\lambda_1$, $\phi_2$, and $\lambda_2$ are the latitudes and longitudes (in radians) of consecutive GPS points. The average speed between two points is then calculated as:

\begin{align}
v &= \frac{d}{\Delta t},
\tag{A2}\label{eq:avg_speed}
\end{align}

where $\Delta t$ represents the elapsed time (in seconds) between successive samples. This real-time computation allows for efficient on-device speed estimation while maintaining privacy, since only scalar speed values—not precise location traces—are required for model input.

These calculations are performed in real time within the application's frontend, allowing users to receive immediate feedback on their trip statistics. Simultaneously, the raw coordinate data, computed speeds, distances, and timestamps are securely transmitted to a cloud database via encrypted API calls. This cloud infrastructure also hosts our pre-trained SpeedTransformer and LSTM-Attention models, enabling real-time transportation mode prediction.

During our experiments, the application's tracking module collected GPS trajectories from a diverse range of smartphone devices and operating systems. Table~\ref{tab:mini_program_summary} summarizes the dataset collected through the mini-program, which exhibits natural variations in sampling frequency and signal quality—characteristics typical of real-world usage. This authentic dataset proved invaluable for validating the SpeedTransformer model under conditions that closely mirror practical deployment environments.

\section{Full Grid Hyper-Parameter Space Search Results \& Analysis}
\label{appendix:gridsearch}

\textbf{Experimental reporting conventions.} 
The following tables enumerate all training and fine-tuning runs conducted in this experiment. To improve readability, we omit dataset size and run identifiers, and instead report the complete set of hyper-parameters that uniquely define each configuration. 
Unless otherwise specified, all accuracies correspond to the test split (\%).

\textbf{Column definitions.} 
\textit{LR}, \textit{BS}, \textit{DO}, \textit{WD}, \textit{Ep.}, and \textit{Early} denote learning rate, batch size, dropout, weight decay, maximum epochs, and early-stopping patience, respectively. 
For the LSTM-Attention model, additional columns include \textit{Hidden} (hidden state dimension) and \textit{Layers} (number of stacked LSTM layers). 
For the SpeedTransformer, columns include \textit{Heads} (attention heads), \textit{d\_model} (embedding dimension), \textit{KV Heads} (shared key/value heads when using grouped-query attention), \textit{Warmup} (number of warmup steps), and \textit{Freeze Policy} (which modules were frozen or reinitialized). 
In fine-tuning tables, \textit{Subset (\%)} indicates the percentage of the target training data used (e.g., 15\%, 20\%, etc.).

\textbf{Summary of results.} 
Tables~\ref{table:lstm-mobis},~\ref{table:transformer-mobis},~\ref{table:lstm-geolife},~\ref{table:transformer-geolife},~\ref{table:LSTM-geolife},~\ref{table:transformer-mobis-geolife},~\ref{table:lstm-mobis-app}, and~\ref{table:transformer-mobis-app} report the results. Across all training and fine-tuning experiments, SpeedTransformer consistently achieved higher accuracy and stronger cross-domain generalization than the LSTM-Attention baseline, particularly under limited-data and transfer learning. When trained directly on the Geolife and MOBIS datasets, both models converged to high accuracy (approximately 93--96\%). However, notable differences emerged in transfer learning and fine-tuning scenarios.

For in-domain training, SpeedTransformer exhibited optimal performance with a learning rate of $2\times10^{-4}$, batch size of 512, dropout of 0.1, and embedding dimension $d_{model}=128$ with 8 attention heads. Accuracy reached $95.97\%$ on Geolife and $94.22\%$ on MOBIS, with stability across modest parameter variations, suggesting strong generalization capacity without overfitting. LSTM-Attention achieved comparable accuracy (around $92.7\%$) using a learning rate of $1\times10^{-3}$, hidden dimension 128--256, and two recurrent layers, but its performance was more sensitive to learning rate changes.

In cross-dataset transfer (MOBIS $\rightarrow$ Geolife), fine-tuned SpeedTransformer models surpassed 85\% test accuracy under optimal configurations, while LSTM-Attention plateaued around 84\%. Among transformer variants, the best results were obtained when the last block was reinitialized ($85.7\%$) or when no layers were frozen ($84.2\%$). 
By contrast, freezing attention or feed-forward layers caused performance to drop below $65\%$, confirming that full end-to-end adaptation is necessary for effective transfer across mobility domains.

Fine-tuning from transfer data (MOBIS $\rightarrow$ Real-World App Experiment) further demonstrated the efficiency of the transformer architecture. Even when using only 15\% of the experiment dataset, SpeedTransformer achieved $89.1\%$ accuracy, outperforming LSTM-Attention ($86.6\%$) under comparable settings. As the fine-tuning subset increased, SpeedTransformer’s accuracy scaled up smoothly, reaching $94.2\%$ with 50\% of the target data. Embedding-freezing strategies yielded intermediate performance (up to $87.8\%$), while partial freezing or warmup schedules offered no clear benefit. These results indicate that SpeedTransformer effectively leverages prior mobility representations with minimal data, whereas LSTM-Attention requires larger sample sizes to reach similar accuracy.

Overall, the experiments reveal three consistent patterns. First, moderate learning rates ($2$--$5\times10^{-4}$) and full-layer fine-tuning yield the most robust convergence across datasets. Second, SpeedTransformer’s attention-based representations exhibit superior transferability and resilience to dataset heterogeneity compared with recurrent encoders, shown in Figure~\ref{fig:transfer}. Third, increasing the fine-tuning subset improves performance approximately monotonically, suggesting that the pre-trained mobility embeddings capture domain-invariant movement structures that generalize across tracking environments.

\newcommand{\compacttablesetup}{%
  \scriptsize
  \setlength{\tabcolsep}{3pt}%
  \renewcommand{\arraystretch}{1.05}%
}

\begin{table}[t]
\compacttablesetup
\caption{Summary of LSTM–Attention performance trained on MOBIS. All runs use Ep.=50, WD=1e-4, EarlyStop=7, unless noted.}
\label{tab:mobis_lstm_summary}
\centering
\begin{adjustbox}{width=\columnwidth}
\begin{tabular}{lcccccc}
\toprule
\textbf{Configuration} & \textbf{LR} & \textbf{Batch Size} & \textbf{Dropout} & \textbf{Hidden Units} & \textbf{Layers} & \textbf{Accuracy (\%)} \\
\midrule
Base (best) & $1\times10^{-3}$ & 128 & 0.1 & 128 & 2 & \textbf{92.40} \\
Smaller batch & $1\times10^{-3}$ & 64 & 0.1 & 128 & 2 & 92.20 \\
Higher dropout & $1\times10^{-3}$ & 64 & 0.2 & 128 & 2 & 92.15 \\
Deeper network & $1\times10^{-3}$ & 64 & 0.1 & 128 & 3 & 92.26 \\
Larger hidden size & $1\times10^{-3}$ & 64 & 0.1 & 256 & 2 & 92.04 \\
Higher learning rate & $2\times10^{-3}$ & 64 & 0.1 & 128 & 2 & 92.14 \\
Lower learning rate & $5\times10^{-4}$ & 128 & 0.1 & 256 & 3 & 92.33 \\
Smaller batch, lower LR & $5\times10^{-4}$ & 64 & 0.1 & 128 & 2 & 92.35 \\
\bottomrule
\end{tabular}
\end{adjustbox}
\label{table:lstm-mobis}
\end{table}

\begin{table}[t]
\compacttablesetup
\caption{Summary of SpeedTransformer performance trained on MOBIS. All runs use Ep.=50, WD=1e-4, EarlyStop=7.}
\label{tab:mobis_speed_summary}
\centering
\begin{adjustbox}{width=\columnwidth}
\begin{tabular}{lccccccc}
\toprule
\textbf{Configuration} & \textbf{LR} & \textbf{Batch Size} & \textbf{Dropout} & \textbf{d\_model} & \textbf{Heads / KV Heads} & \textbf{Accuracy (\%)} \\
\midrule
Base (best) & $1\times10^{-4}$ & 512 & 0.1 & 128 & 8 / 4 & \textbf{94.22} \\
Higher LR & $2\times10^{-4}$ & 512 & 0.1 & 128 & 8 / 2 & 94.20 \\
Larger model & $2\times10^{-4}$ & 512 & 0.1 & 192 & 12 / 6 & 93.71 \\
Larger embedding & $2\times10^{-4}$ & 512 & 0.1 & 256 & 8 / 4 & 93.54 \\
Smaller batch & $2\times10^{-4}$ & 1024 & 0.1 & 128 & 8 / 4 & 93.69 \\
Higher dropout & $2\times10^{-4}$ & 512 & 0.2 & 128 & 8 / 4 & 93.54 \\
Higher LR & $3\times10^{-4}$ & 512 & 0.1 & 128 & 8 / 4 & 94.09 \\
\bottomrule
\end{tabular}
\end{adjustbox}
\label{table:transformer-mobis}
\end{table}

\begin{table}[t]
\compacttablesetup
\caption{Summary of LSTM–Attention performance trained on Geolife. All runs use Ep.=50, WD=1e-4, EarlyStop=7, unless noted.}
\label{tab:gl_lstm_summary}
\centering
\begin{adjustbox}{width=\columnwidth}
\begin{tabular}{lcccccc}
\toprule
\textbf{Configuration} & \textbf{LR} & \textbf{Batch Size} & \textbf{Dropout} & \textbf{Hidden Units} & \textbf{Layers} & \textbf{Accuracy (\%)} \\
\midrule
Base (best) & $1\times10^{-3}$ & 128 & 0.1 & 128 & 2 & \textbf{92.77} \\
Smaller batch & $1\times10^{-3}$ & 64 & 0.1 & 128 & 2 & 91.77 \\
Higher dropout & $1\times10^{-3}$ & 64 & 0.2 & 128 & 2 & 92.71 \\
Deeper network & $1\times10^{-3}$ & 64 & 0.1 & 128 & 3 & 92.16 \\
Larger hidden size & $1\times10^{-3}$ & 64 & 0.1 & 256 & 2 & 92.73 \\
Higher learning rate & $2\times10^{-3}$ & 64 & 0.1 & 128 & 2 & 92.56 \\
Lower learning rate & $5\times10^{-4}$ & 128 & 0.1 & 256 & 3 & 92.40 \\
Smaller batch, lower LR & $5\times10^{-4}$ & 64 & 0.1 & 128 & 2 & 91.63 \\
\bottomrule
\end{tabular}
\end{adjustbox}
\label{table:lstm-geolife}
\end{table}

\begin{table}[t]
\compacttablesetup
\caption{Summary of SpeedTransformer performance trained on Geolife. All models use Ep.=50, WD=1e-4, EarlyStop=7, unless noted.}
\label{tab:gl_summary}
\centering
\begin{adjustbox}{width=\columnwidth}
\begin{tabular}{lcccc}
\toprule
\textbf{Configuration} & \textbf{Learning Rate (LR)} & \textbf{Model Size ($d_\text{model}$)} & \textbf{Attention Heads (Q/KV)} & \textbf{Accuracy (\%)} \\
\midrule
Base (optimal) & $2\times10^{-4}$ & 128 & 8 / 4 & \textbf{95.97} \\
Reduced KV heads & $2\times10^{-4}$ & 128 & 8 / 2 & 95.36 \\
Larger model & $2\times10^{-4}$ & 192 & 12 / 6 & 94.72 \\
Larger embedding & $2\times10^{-4}$ & 256 & 8 / 4 & 94.69 \\
Higher LR & $3\times10^{-4}$ & 128 & 8 / 4 & 94.68 \\
Lower LR & $1\times10^{-4}$ & 128 & 8 / 4 & 94.44 \\
Higher batch size & $2\times10^{-4}$ & 128 & 8 / 4 & 92.72 \\
Higher dropout & $2\times10^{-4}$ & 128 & 8 / 4 & 92.70 \\
\bottomrule
\end{tabular}
\end{adjustbox}
\label{table:transformer-geolife}
\end{table}

\begin{table}[t]
\compacttablesetup
\caption{Summary of fine-tuning LSTM-Attention (MOBIS $\rightarrow$ Geolife). All models use BS=128, DO=0.3, WD=1e-4, Ep.=60, and EarlyStop=7.}
\label{tab:ft_summary_lstm}
\centering
\begin{adjustbox}{width=\columnwidth}
\begin{tabular}{lccc}
\toprule
\textbf{Learning Rate (LR)} & \textbf{Hidden Size} & \textbf{Accuracy (\%)} & \textbf{Notes} \\
\midrule
$1\times10^{-3}$ & 64–256 & \textbf{84.17} & Optimal configuration \\
$5\times10^{-4}$ & 64–256 & 84.13 & Comparable to best \\
$1\times10^{-4}$ & 64–256 & 82.84 & Slight underfitting \\
$5\times10^{-5}$ & — & 75.47–79.15 & Low performance (few epochs) \\
\bottomrule
\end{tabular}
\end{adjustbox}
\label{table:LSTM-geolife}
\end{table}

\begin{table}[t]
\compacttablesetup
\caption{Summary of fine-tuning SpeedTransformer (MOBIS $\rightarrow$ Geolife). All models use BS=512, WD=1e-4, DO=0.2, Ep.=50, and EarlyStop=7.}
\label{tab:ft_summary}
\centering
\begin{adjustbox}{width=\columnwidth}
\begin{tabular}{lccc}
\toprule
\textbf{Learning Rate (LR)} & \textbf{Best Freeze Policy} & \textbf{Warmup Steps} & \textbf{Accuracy (\%)} \\
\midrule
$5\times10^{-5}$ & Reinit last block & 0 & 85.07 \\
$1\times10^{-4}$ & Reinit last block & 0 & 85.70 \\
$2\times10^{-4}$ & Attention frozen  & 0 & \textbf{86.38} \\
$2\times10^{-4}$ & Reinit last block & 0 & 86.30 \\
\midrule
$1\times10^{-4}$ & Embeddings frozen & 0 & 84.22 \\
$1\times10^{-4}$ & None (all trainable) & 0 & 84.08 \\
$2\times10^{-4}$ & None (all trainable) & 0 & 85.01 \\
\midrule
$5\times10^{-5}$ to $2\times10^{-4}$ & Any freeze w/ warmup (100–500) & — & 63–67 (no gain) \\
\bottomrule
\end{tabular}
\end{adjustbox}
\label{table:transformer-mobis-geolife}
\end{table}

\begin{table}[t]
\compacttablesetup
\caption{Summary of LSTM-Attention fine-tuning results from MOBIS $\rightarrow$ Real-World App Experiment. All runs use 50 epochs, EarlyStop with patience = 10 (Pat.\,10), dropout = 0.3, and weight decay = 1e-4.}
\label{tab:ft_m2mini_lstm_summary}
\centering
\begin{adjustbox}{width=\columnwidth}
\begin{tabular}{lcccccc}
\toprule
\textbf{Subset (\%)} & \textbf{LR} & \textbf{Hidden Units} & \textbf{Batch Size} & \textbf{Early Stop} & \textbf{Pat.} & \textbf{Accuracy (\%)} \\
\midrule
15 & $5\times10^{-4}$ & 128 & 128 & 10 & Pat.10 & 86.62 \\
20 & $5\times10^{-4}$ & 128 & 128 & 10 & Pat.10 & 88.02 \\
30 & $5\times10^{-4}$ & 128 & 128 & 10 & Pat.10 & \textbf{88.97} \\
40 & $5\times10^{-4}$ & 128 & 128 & 10 & Pat.10 & 87.45 \\
50 & $5\times10^{-4}$ & 128 & 128 & 10 & Pat.10 & 87.57 \\
\midrule
15 & $1\times10^{-3}$ & 128--256 & 128 & 10 & Pat.10 & 86.19 \\
15 & $1\times10^{-4}$ & 128--256 & 128 & 10 & Pat.10 & 85.18 \\
\bottomrule
\end{tabular}
\end{adjustbox}
\label{table:lstm-mobis-app}
\end{table}

\begin{table}[t]
\compacttablesetup
\caption{Summary of SpeedTransformer fine-tuning results from MOBIS $\rightarrow$ Real-World App Experiment. All runs used 50 epochs, EarlyStop with patience = 10 (Pat.\,10),  dropout = 0.2, and weight decay = 1e-4.}
\label{tab:ft_m2mini_tr_summary}
\centering
\begin{adjustbox}{width=\columnwidth}
\begin{tabular}{lcccccc}
\toprule
\textbf{Subset (\%)} & \textbf{Learning Rate} & \textbf{Warmup Steps} & \textbf{Freeze Policy} & \textbf{Batch Size} & \textbf{Early Stop} & \textbf{Accuracy (\%)} \\
\midrule
15 & $5\times10^{-4}$ & 0 & None & 512 & Pat.10 & 89.12 \\
20 & $5\times10^{-4}$ & 0 & None & 512 & Pat.10 & 82.53 \\
30 & $5\times10^{-4}$ & 0 & None & 512 & Pat.10 & \underline{91.15} \\
40 & $5\times10^{-4}$ & 0 & None & 512 & Pat.10 & 88.78 \\
50 & $5\times10^{-4}$ & 0 & None & 512 & Pat.10 & \textbf{94.22} \\
\midrule
15 & $1\times10^{-4}$ & 0--100 & Attention/Embedding frozen & 512 & Pat.10 & 69.3–72.9 \\
15 & $2\times10^{-4}$ & 0--100 & Embeddings frozen           & 512 & Pat.10 & 71.6–80.6 \\
15 & $5\times10^{-4}$ & 0--100 & Embeddings frozen           & 512 & Pat.10 & 77.2–87.8 \\
\bottomrule
\end{tabular}
\end{adjustbox}
\label{table:transformer-mobis-app}
\end{table}

\section{Input Embeddings}
\label{appendix:embedding}
Each scalar speed value \( s_t \) is linearly projected into a \( d_{model} = 128 \)-dimensional embedding space, producing an input embedding matrix \( \mathbf{E} \in \mathbb{R}^{T \times d_{model}} \). This projection enables the model to represent one-dimensional scalar speeds within a high-dimensional latent space suitable for Transformer-based sequence modeling:

\begin{equation}
\mathbf{e}_t = \mathbf{W}_e s_t + \mathbf{b}_e,
\end{equation}

where \( s_t \) denotes the scalar speed at time step \( t \), \( \mathbf{W}_e \in \mathbb{R}^{1 \times d} \) is the learnable weight matrix, and \( \mathbf{b}_e \in \mathbb{R}^{d} \) is the learnable bias vector. The resulting embeddings \( \mathbf{E} = [\mathbf{e}_1, \mathbf{e}_2, \ldots, \mathbf{e}_T] \) serve as the model’s input sequence to subsequent positional encoding and self-attention layers.

\section{SwiGLU Activation}
\label{swiglu}
Each feed-forward subnetwork adopts the SwiGLU activation \citep{Shazeer2020GLU}, defined as:
\begin{equation}
\text{SwiGLU}(\mathbf{x}) = (\mathbf{x}\mathbf{W}_1) \odot \text{Swish}(\mathbf{x}\mathbf{W}_2)\mathbf{W}_3,
\end{equation}
where \( \odot \) denotes element-wise multiplication, \( \mathbf{W}_1, \mathbf{W}_2, \mathbf{W}_3 \) are learnable weight matrices, and \(\text{Swish}(\mathbf{x}) = \mathbf{x}\sigma(\mathbf{x})\) with \(\sigma(\cdot)\) being the sigmoid activation function. SwiGLU enhances representational expressivity and improves training stability relative to standard ReLU-based feed-forward layers.

\section{Model Window Size}
\label{window_size}

\begin{figure}[ht]
    \centering
    \includegraphics[width=0.8\textwidth]{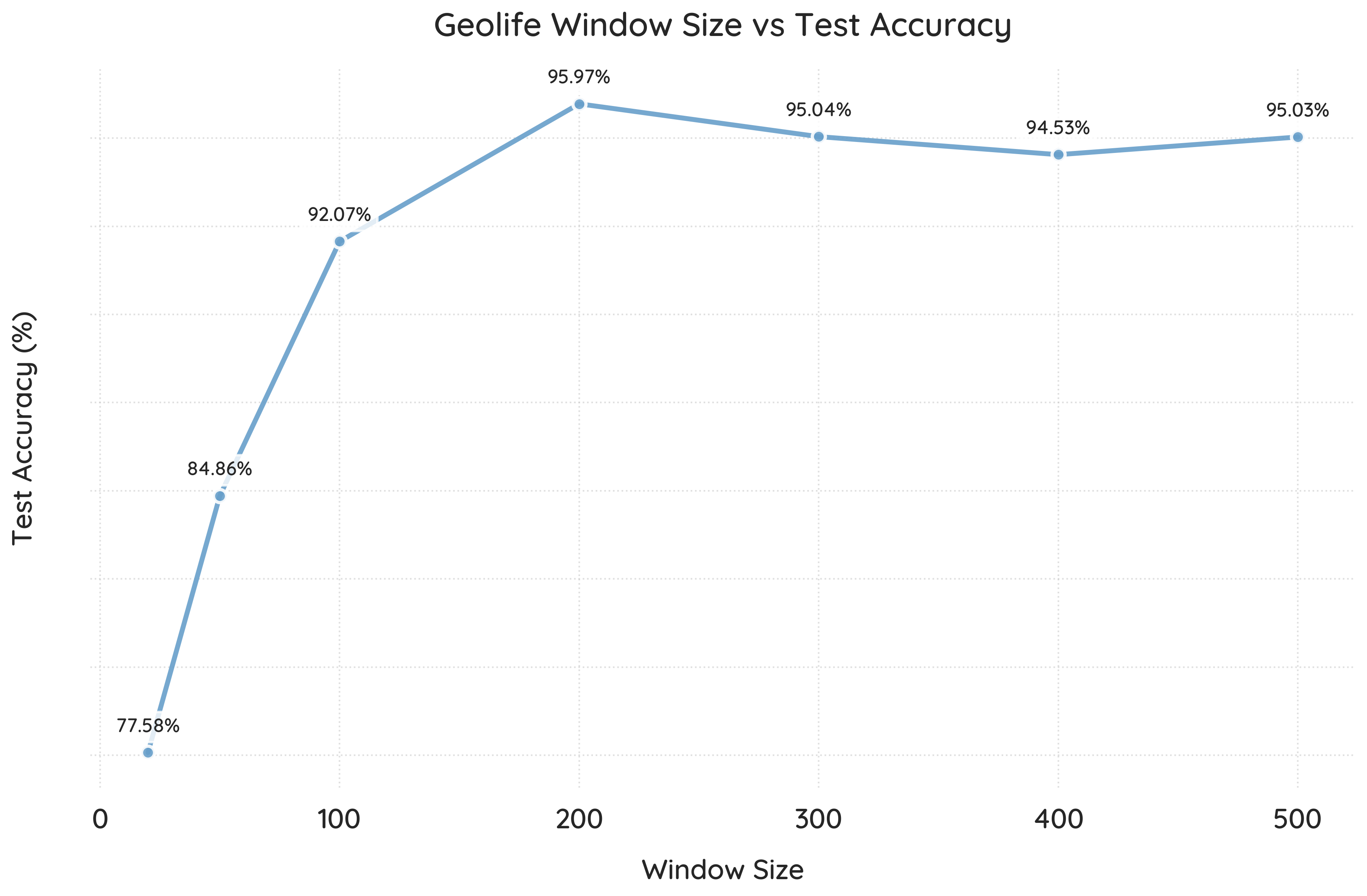}
    \caption{Effect of window size on test accuracy for the Geolife dataset. A window size of 200 provides the optimal balance between temporal context and computational efficiency, achieving the highest accuracy (95.97\%).}
    \label{fig:window_size_accuracy}
\end{figure}

\bigskip
\noindent
Figure~\ref{fig:window_size_accuracy} supports our choice of $T=200$, showing that performance improves steadily with larger window sizes up to 200, beyond which accuracy plateaus and slightly declines. The 500-sample configuration also required a reduced batch size due to GPU memory constraints, potentially impacting training stability. Overall, $T=200$ represents the optional size between accuracy, computational efficiency, and training stability, capturing sufficient motion dynamics without introducing unnecessary redundancy or overfitting.

\begin{table}[htbp]
\centering
\caption{Test accuracy and loss across different window sizes (Geolife dataset).}
\label{tab:winlog}
\begin{tabular}{lcccc}
\hline
\textbf{W. Size} & \textbf{Test Acc. (\%)} & \textbf{Test Loss} & \textbf{Run Name} \\
\hline
20  & 77.58 & 0.5778 & Geolife\_ws20\_lr2e-4\_bs512\_h8\_d128\_kv4\_do0.1 \\
50  & 84.86 & 0.4043 & Geolife\_ws50\_lr2e-4\_bs512\_h8\_d128\_kv4\_do0.1 \\
100 & 92.07 & 0.2600 & Geolife\_ws100\_lr2e-4\_bs512\_h8\_d128\_kv4\_do0.1 \\
200 & 95.97 & 0.1525 & Geolife\_ws200\_lr2e-4\_bs512\_h8\_d128\_kv4\_do0.1 \\
300 & 95.04 & 0.1753 & Geolife\_ws300\_lr2e-4\_bs512\_h8\_d128\_kv4\_do0.1 \\
400 & 94.53 & 0.1703 & Geolife\_ws400\_lr2e-4\_bs256\_h8\_d128\_kv4\_do0.1 \\
500 & 95.03 & 0.1800 & Geolife\_ws500\_lr2e-4\_bs256\_h8\_d128\_kv4\_do0.1 \\
\hline
\end{tabular}
\end{table}

\section{SpeedTransformer in Original Transformer Model Architecture}
\label{appendix:old-transformer}

We also conduct experiments using the original Transformer architecture, proposed by \citet{vaswani_attention_2017}, for a comparison. Although the core architecture shares the transformer encoder backbone with the final model presented in Section~\ref{sec:arch}, this initial version employed sinusoidal positional encodings and a standard feed-forward block instead of Rotary Positional Embeddings (RoPE) and SwiGLU activations. The updated design described in the main text improved both training stability, computational efficiency, and cross-dataset transferability.

The following subsections summarize the original architecture, input encoding strategy, and optimization setup for the model. We provide experiment results, which serves as a reference for ablation comparisons.

\begin{figure}[ht]
    \centering
    \includegraphics[width=1\textwidth]{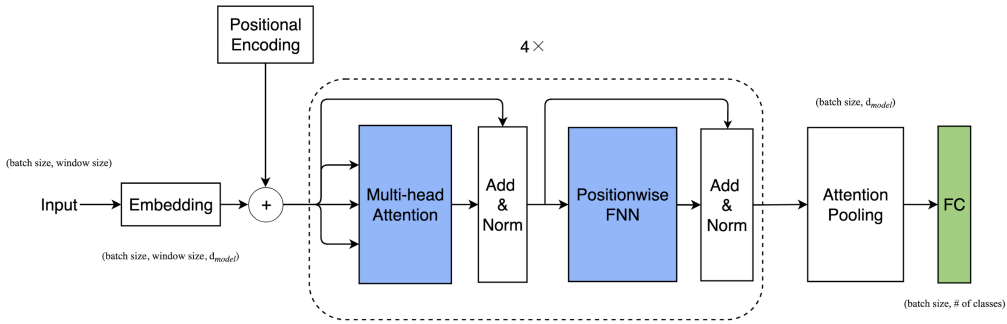}
    \caption{Transformer architecture for transportation mode classification.}
    \label{fig:workflow}
\end{figure}

The model architecture adapts the transformer encoder framework \citep{vaswani_attention_2017}. Figure \ref{fig:workflow} presents the overall architecture of our model. To incorporate sequential order information, we add sinusoidal positional encoding to these embeddings, shown in Equation~\ref{eq:sinusoidal}.

\begin{align}
\mathbf{PE}{(pos, 2i)} &= \sin\Bigl(\frac{pos}{10000^{2i/d_{\text{model}}}}\Bigr), \quad
\mathbf{PE}{(pos, 2i+1)} = \cos\Bigl(\frac{pos}{10000^{2i/d_{\text{model}}}}\Bigr)
\label{eq:sinusoidal}
\end{align}

where $pos$ is the position in the sequence, $i$ is the dimension index, and $d_{\text{model}}$ is the embedding dimension.

We use the same 200 variable-length sliding window, in consistent with the main model. We projects scalar speed value into high-dimensional vector representations through linear transformation followed by non-linear activation in Equation~\ref{eq: relu}.

\begin{equation}
\mathbf{E} = \text{ReLU}(\mathbf{W}_e \cdot \mathbf{S} + \mathbf{b}_e)
\label{eq: relu}
\end{equation}
where $\mathbf{S}$ represents input speed values and $\mathbf{E}$ the resulting embeddings. 

These encoding enables the model to differentiate positions and captures temporal variations that are critical to distinguish transportation modes over time. For example, information such as acceleration patterns can be inferred from sequences of trajectories.

\begin{figure}[tb]
\centering
\begin{tikzpicture}[x=1.6cm,y=1.1cm,
  every node/.style={font=\small}]
  \node[anchor=west,font=\bfseries] at (0,0) {Trajectory A (5 Hz samples)};
  \foreach \i/\val in {0/4.2,1/7.8,2/12.1,3/7.8,4/3.0}{
    \node[draw, rounded corners=2pt, minimum width=1.2cm, minimum height=.6cm,
          anchor=west] (A\i) at (\i,-0.8) {\val};
  }
  \draw[very thick,rounded corners=2pt,purple]
    ([xshift=-.06cm,yshift=-.06cm]A1.north west) rectangle
    ([xshift=.06cm,yshift=.06cm]A1.south east);
  \draw[very thick,rounded corners=2pt,purple]
    ([xshift=-.06cm,yshift=-.06cm]A3.north west) rectangle
    ([xshift=.06cm,yshift=.06cm]A3.south east);

  \node[anchor=west,font=\bfseries] at (0,-2.2)
        {Trajectory B (same speeds, different order)};
  \foreach \i/\val in {0/7.8,1/4.2,2/3.0,3/12.1,4/7.8}{
    \node[draw, rounded corners=2pt, minimum width=1.2cm, minimum height=.6cm,
          anchor=west] (B\i) at (\i,-3.0) {\val};
  }
  \draw[very thick,rounded corners=2pt,orange]
    ([xshift=-.06cm,yshift=-.06cm]B0.north west) rectangle
    ([xshift=.06cm,yshift=.06cm]B0.south east);
  \draw[very thick,rounded corners=2pt,orange]
    ([xshift=-.06cm,yshift=-.06cm]B4.north west) rectangle
    ([xshift=.06cm,yshift=.06cm]B4.south east);
\end{tikzpicture}
\caption{Two GPS speed sequences share the same multiset of speeds but in different orders. Without position information the embeddings for repeated values (e.g., \(7.8\,m/s\)) are identical. With positional encoding we form \(\mathbf{z}_t=E(v_t)+P(t)\), allowing attention to model order-dependent patterns.}
\label{fig:pos-vel-encoding}
\end{figure}
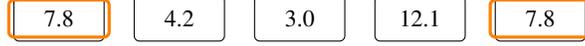

After the positional encoding, the inputs are fed into a self-attention layer, which helps the encoder to check speed at other positions in the input sequence as it encodes a speed at a specific location, as shown in Figure ~\ref{fig:pos-vel-encoding}. \citet{vaswani_attention_2017} established that attention mechanisms compute weighted aggregations of value vectors where weights are determined by compatibility scores between query and key vectors, which we adopt the scaled dot-product attention formulation as defined in Equation (1) of Vaswani et al. (2017) and use their definition of scaled dot-product attention in Equation~\ref{eq:attention}.
\begin{equation}
\text{Attention}(\mathbf{Q}, \mathbf{K}, \mathbf{V}) = \text{softmax}\Bigl(\frac{\mathbf{Q}\mathbf{K}^\top}{\sqrt{d_k}}\Bigr)\mathbf{V}
\label{eq:attention}
\end{equation}

In this formulation, $\mathbf{Q}$ represents queries that seek information, $\mathbf{K}$ encodes keys that store information, and $\mathbf{V}$ contains values that are aggregated according to query-key compatibility. For self-attention, all three matrices derive from the same input sequence, which enables each position to attend to all positions within the sequence.

The multi-head attention mechanism extends this concept by computing attention in parallel across different representation subspaces. We follow \citet{vaswani_attention_2017} and use their definition of multihead representation in Equation~\ref{eq:multihead}.
\begin{equation}
\text{MultiHead}(\mathbf{Q}, \mathbf{K}, \mathbf{V}) = \text{Concat}(\text{head}1, \ldots, \text{head}_h)\mathbf{W}^O
\end{equation}
\begin{equation}
\text{head}_i = \text{Attention}(\mathbf{Q}\mathbf{W}^Q_i, \mathbf{K}\mathbf{W}^K_i, \mathbf{V}\mathbf{W}^V_i)
\label{eq:multihead}
\end{equation}

Here, $\mathbf{W}^Q_i$, $\mathbf{W}^K_i$, and $\mathbf{W}^V_i$ are learned projection matrices that transform the original embeddings into different subspaces, while $\mathbf{W}^O$ projects the concatenated outputs back to the model dimension.

In line with standard deep learning practice, each encoder layer includes a position-wise feed-forward network (FFN) applied independently to every position. The input to the FFN is formed by summing the token embeddings and positional encodings, as shown in Equation~\ref{eq:ffn_input}:

\begin{equation}
    \mathbf{z} = \mathbf{x} + \mathbf{PE},
    \label{eq:ffn_input}
\end{equation}
where $\mathbf{x}$ denotes the token embedding and $\mathbf{PE}$ represents the positional encoding.

Residual connections and layer normalization are applied around each sub-layer to facilitate gradient flow and enhance training stability.  
To generate a fixed-length representation from variable-length sequences, we employ attention-based pooling. The position-aware input $\mathbf{z}$ is then passed into the feed-forward network (FFN), defined in Equation~\ref{eq:ffn}:

\begin{equation}
    \text{FFN}(\mathbf{x} + \mathbf{PE}) = \max\left(0, (\mathbf{x} + \mathbf{PE}) \mathbf{W}_1 + \mathbf{b}_1\right)\mathbf{W}_2 + \mathbf{b}_2,
    \label{eq:ffn}
\end{equation}

where $\mathbf{W}_1$ and $\mathbf{W}_2$ are weight matrices, and $\mathbf{b}_1$ and $\mathbf{b}_2$ are bias terms.  
To incorporate sequential order information, token embeddings are first augmented with sinusoidal positional encoding through the attention mechanism and softmax normalization. These position-aware inputs are then processed by the FFN in Equation~\ref{eq:ffn}, enabling the model to capture complex non-linear transformations of input representations at each position.

The model is trained using the cross-entropy loss function, which quantifies the divergence between the predicted and ground-truth transportation modes, as shown in Equation~\ref{eq:cross_entropy}:

\begin{equation}
\mathcal{L} = -\sum_{i=1}^{N} \sum_{c=1}^{C} y_{i,c} \log(p_{i,c}),
\label{eq:cross_entropy}
\end{equation}

\noindent
where $N$ is the number of samples, $C$ is the number of classes, $y_{i,c} \in \{0, 1\}$ denotes the ground-truth label (1 if sample $i$ belongs to class $c$), and $p_{i,c}$ is the predicted probability of class $c$ for sample $i$.

Final classification is performed through a linear projection of the pooled representation, followed by a softmax activation to generate a probability distribution over transportation modes, as defined in Equation~\ref{eq:softmax}:

\begin{equation}
p(y=c \mid \mathbf{c}) = \frac{\exp(\mathbf{w}_c^\top \mathbf{c} + b_c)}{\sum_{j=1}^{C} \exp(\mathbf{w}_j^\top \mathbf{c} + b_j)},
\label{eq:softmax}
\end{equation}

\noindent
where $\mathbf{c}$ denotes the pooled feature vector, and $\mathbf{w}_c$ and $b_c$ represent the weight vector and bias term for class $c$, respectively.

We implement AdamW optimization with weight decay, learning rate scheduling, dropout regularization, and gradient clipping. This architecture enables efficient discovery of complex temporal patterns in speed data, capturing the distinctive signatures of different transportation modes without requiring additional input features or pre-processing steps.


\begin{table}[ht]
\centering
\caption{Per-class accuracy (recall) of LSTM and Transformer models across datasets. 
Each model was trained on Geolife and MOBIS, and fine-tuned from MOBIS to Geolife and Mini-Program datasets.}
\label{tab:old_results}
\begin{tabular}{llccccc}
\toprule
\textbf{Model} & \textbf{Training Setup} & \textbf{Bike} & \textbf{Bus} & \textbf{Car} & \textbf{Train} & \textbf{Walk} \\
\midrule
\multirow{4}{*}{LSTM}
& Geolife                & 0.92 & 0.86 & 0.86 & 0.92 & 0.99 \\
& MOBIS                  & 0.61 & 0.84 & 0.99 & 0.28 & 0.99 \\
& MOBIS $\rightarrow$ Geolife     & 0.59 & 0.84 & 0.57 & 0.79 & 0.72 \\
& MOBIS $\rightarrow$ Mini-Program & 0.39 & 0.69 & 0.93 & 0.79 & 0.86 \\
\midrule
\multirow{4}{*}{Transformer}
& Geolife                & 0.93 & 0.96 & 0.92 & 0.87 & 0.99 \\
& MOBIS                  & 0.78 & 0.88 & 0.98 & 0.43 & 0.99 \\
& MOBIS $\rightarrow$ Geolife     & 0.75 & 0.88 & 0.75 & 0.80 & 0.99 \\
& MOBIS $\rightarrow$ Mini-Program & 0.40 & 0.70 & 1.00 & 0.43 & 0.99 \\
\bottomrule
\end{tabular}
\end{table}

As shown in Table~\ref{tab:old_results}, the Transformer consistently outperforms the LSTM across all datasets and transfer setups, confirming its superior ability to capture temporal dependencies and generalize across domains.

\section{Rule-based Model}
\label{appendix:rule-based}

Several prior studies on transportation mode detection rely on simple rule-based models built on handcrafted heuristics derived from dense GPS trajectory data \citep{huang2019transport}. Such models typically depend on a small number of summary statistics (e.g., speed percentiles, stop ratios, or acceleration variability), require no auxiliary GIS data, and are computationally inexpensive. However, their performance is highly sensitive to threshold choices and to behavioral heterogeneity across users, cities, and datasets.

To establish a strong heuristic baseline, we implemented a hierarchical rule-based classifier operating on sliding windows of raw GPS trajectories. Unlike earlier implementations that rely exclusively on fixed expert-defined thresholds, we calibrated the rule parameters separately for each dataset using the training split, yielding dataset-specific but still purely heuristic decision rules. The rules operate as follows. First, the 95th-percentile speed within each window is used to separate low-speed modes (walk, bike) from motorized and rail modes. Second, windows exceeding a high-speed threshold are classified as rail. Finally, among remaining motorized windows, the stop ratio and acceleration variability are used to distinguish buses from cars.

Table~\ref{tab:heuristic-rule} summarizes the calibrated thresholds used in our experiments. All speed values are expressed in meters per second (m/s).

\begin{table}[h!]
\centering
\begin{tabular}{p{3.5cm} p{3cm} p{8cm}}
\toprule
\textbf{Rule} & \textbf{Threshold (m/s)} & \textbf{Description / Condition for Mode Assignment} \\
\midrule
\texttt{walk\_p95\_max} & $\leq 3.0$ & Windows with 95th-percentile speed below 3.0\,m/s are classified as \textit{walk}. \\
\texttt{bike\_p95\_max} & $\leq 8.0$ & Windows with 95th-percentile speed between 3.0–8.0\,m/s are classified as \textit{bike}. \\
\texttt{road\_p95\_min} & $\geq 17.4$ (18.0*) & Windows exceeding this threshold are treated as motorized road transport (\textit{bus} or \textit{car}). \\
\texttt{rail\_p95\_min} & $\geq 27.5$ (40.0*) & Windows with very high 95th-percentile speed are classified as \textit{rail}. \\
\texttt{stop\_thresh} & $< 0.5$ & Speeds below 0.5\,m/s are considered ``stopped'' for computing stop ratio. \\
\texttt{bus\_stop\_ratio\_min} & $\geq 0.05$ & Within the motorized range, higher stop ratios indicate \textit{bus}. \\
\texttt{accel\_std\_split} & $< 2.0$ & Smoother acceleration patterns indicate \textit{bus}; higher variability indicates \textit{car}. \\
\bottomrule
\end{tabular}
\caption{Calibrated heuristic rules for transportation mode classification. Thresholds are tuned separately for each dataset using the training split. \newline
*A different threshold was used for Geolife}
\label{tab:heuristic-rule}
\end{table}

We evaluated the rule-based model on both the Geolife and MOBIS datasets using the same train--test splits employed for all learning-based models in the main text. The results are reported in Table~\ref{tab:rule-based_results}. Despite dataset-specific calibration, the rule-based model achieves only moderate overall accuracy on both datasets (0.3967), with substantial variation in per-class performance.

On the MOBIS dataset, the model exhibits high precision but low recall for the dominant \textit{car} class, indicating that while many predicted car windows are correct, a large fraction of true car instances are misclassified as other modes. Conversely, \textit{walk}, \textit{bike}, and \textit{train} achieve relatively high recall but low precision, reflecting extensive confusion driven by overlapping speed regimes and heterogeneous travel behavior. As a result, the weighted F1-score remains modest (0.4929) despite strong class imbalance.

On the Geolife dataset, the rule-based model performs slightly more evenly across classes but still struggles to distinguish between motorized road modes. In particular, \textit{bus} is almost never correctly identified, while \textit{train} and \textit{walk} benefit from their more distinctive speed profiles. These results highlight a fundamental limitation of heuristic approaches: even when carefully calibrated, fixed rules cannot adequately capture the diversity of urban mobility patterns present in real-world GPS data.

Overall, these findings reinforce our broader argument. While rule-based models are transparent, interpretable, and computationally efficient, they are inherently rigid and brittle. In settings where large-scale trajectory data are available and mobility behavior is complex, data-driven models that learn representations directly from raw trajectories offer a far more robust and scalable solution.

\begin{table}[h!]
\centering
\begin{tabular}{lcccccc}
\toprule
\textbf{Dataset} & \textbf{Mode} & \textbf{Precision} & \textbf{Recall} & \textbf{F1-score} & \textbf{Support} & \textbf{Accuracy} \\
\midrule
\multirow{6}{*}{\textbf{Geolife}} 
 & Bike  & 0.2818 & 0.3584 & 0.3155 & 3{,}200 &  \\
 & Bus   & 1.0000 & 0.0000 & 0.0000 & 4{,}863 &  \\
 & Car   & 0.0530 & 0.0871 & 0.0659 & 2{,}251 &  \\
 & Train & 0.4256 & 0.9512 & 0.5881 & 4{,}119 &  \\
 & Walk  & 0.9463 & 0.4856 & 0.6418 & 5{,}225 &  \\
 & \textbf{Macro Avg.} & 0.5413 & 0.3765 & 0.3223 &  &  \\
 & \textbf{Weighted Avg.} & 0.6400 & 0.3967 & 0.3527 &  &  \\
 & \textbf{Overall Accuracy} &  &  &  & 19{,}658 & \textbf{0.3967} \\
\midrule
\multirow{6}{*}{\textbf{MOBIS}} 
 & Bike  & 0.1087 & 0.6552 & 0.1864 & 17{,}389 &  \\
 & Bus   & 0.1428 & 0.3550 & 0.2037 & 14{,}474 &  \\
 & Car   & 0.9301 & 0.3278 & 0.4847 & 336{,}860 &  \\
 & Train & 0.0911 & 0.7243 & 0.1618 & 27{,}607 &  \\
 & Walk  & 0.9886 & 0.4627 & 0.6303 & 155{,}581 &  \\
 & \textbf{Macro Avg.} & 0.4523 & 0.5050 & 0.3334 &  &  \\
 & \textbf{Weighted Avg.} & 0.8581 & 0.3967 & 0.4929 &  &  \\
 & \textbf{Overall Accuracy} &  &  &  & 551{,}911 & \textbf{0.3967} \\
\bottomrule
\end{tabular}
\caption{Performance of the calibrated rule-based model on Geolife and MOBIS datasets.}
\label{tab:rule-based_results}
\end{table}

\section{SpeedTransformer's Computational Efficiency}
\label{appendix:computational_efficiency}

Like most deep learning architectures, SpeedTransformer requires substantially higher computational resources than traditional, non–deep learning models. Nevertheless, through our adaptation of the Grouped-Query Attention (GQA) structure, the model achieves comparatively high computational efficiency.

\begin{figure}[ht]
    \centering
    \includegraphics[width=0.85\linewidth]{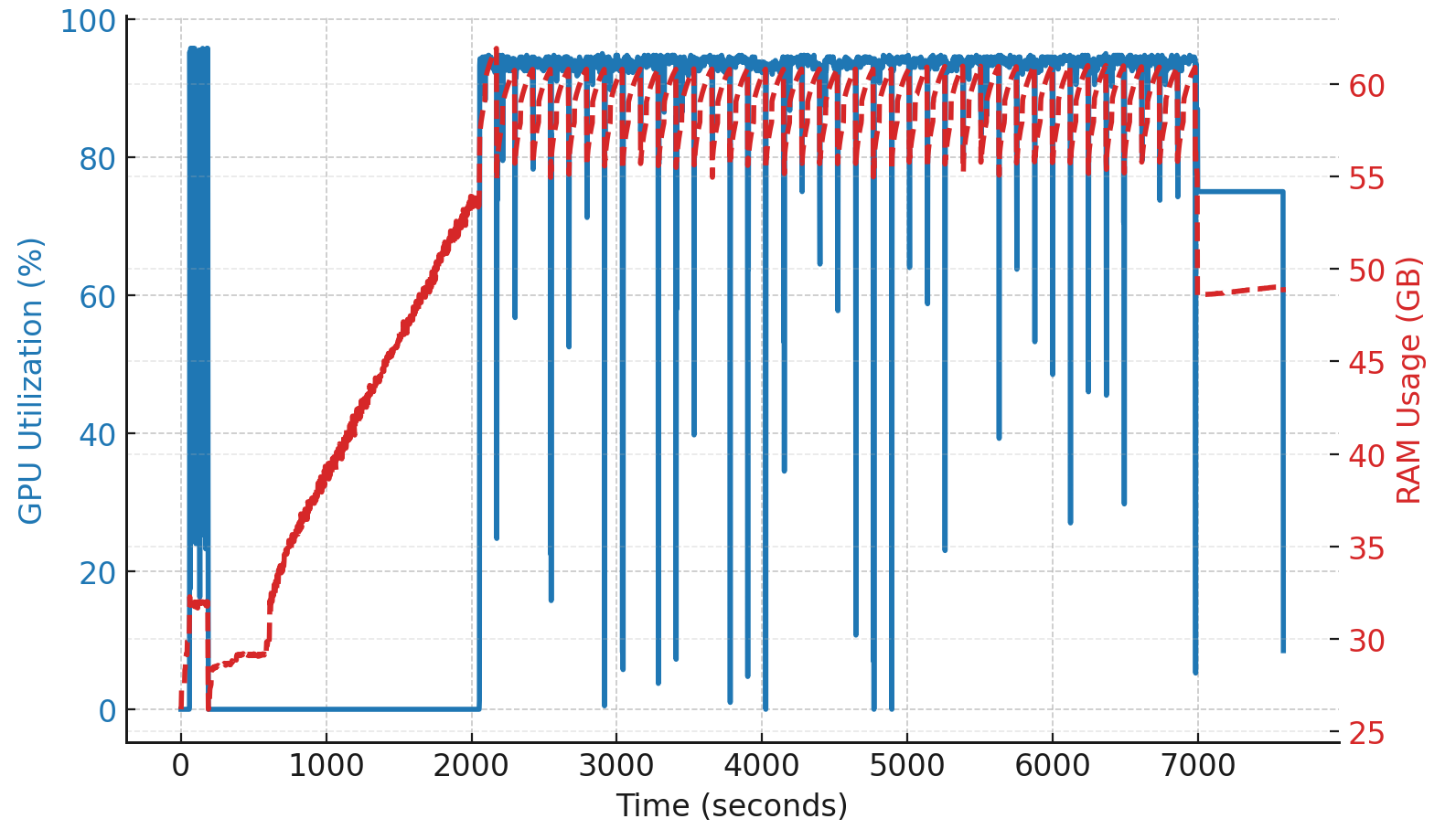}
    \vspace{-2mm}
    \caption{GPU and memory utilization of the Transformer model during training.}
    \label{fig:GPU_memory}
\end{figure}

As illustrated in Figure~\ref{fig:GPU_memory}, during a full training session with four processes one a single A100 GPU node, the model maintained an average GPU utilization of approximately 67\% (peaking at 96\%) across all devices. Meanwhile, host memory usage averaged around 52.9 GB (with a peak of 61.9 GB) over a training period of roughly 126.5 minutes. These measurements suggest effective multi-GPU-process utilization and stable memory management, consistent with a compute-bound workload characteristic of Transformer-based architectures.

\section{IRB Protocol and Ethics Approval}
\label{irb}

Following the Committee on Publication Ethics (COPE) guidance, we include our IRB protocol and ethical statement here, concerning our real-world experiment. In respect for conciseness, we summarize our IRB protocol and its approval below. 

This study was approved by the Institutional Review Board (IRB) at Duke Kunshan University (DKU) shortly before the carried out experiment. The protocol (version dated June 23, 2022) received clearance to conduct randomized experiments using a WeChat-based mini-program developed by the research team. Participants were recruited from cities in China, through both online advertisements and QR-code posters in focus groups.

Eligible participants (aged 18 and above) provided brief informed consent, consistent with standard practices in the industry, before engaging in the study. The consent process was integrated into the mini-program and clearly stated that no personal identifiers (e.g., name, phone number, WeChat ID) would be collected. Instead, a pseudonymous device ID was used to generate a non-identifiable case ID for analysis. Demographic information is purposefully not collected to protect human subjects' privacy. Minimal geographic and behavioral data were collected for the purpose of estimating participants’ daily carbon footprints.

The study involved daily interaction with the mini-program over a one-month period from November 2023 to December 2023, during which participants received varying forms of informational stimulus—ranging from government policy content to scientific facts and social cues. Weekly questionnaires measured participants’ willingness to pay for carbon reduction.

Data were stored securely on DKU-managed servers, and all members of the research team were Chinese citizens, in compliance with China's data sovereignty regulations. No audio, video, or photographic data were collected. Participants had the option to donate or receive a small monetary compensation for their time.

This research posed minimal risk to participants and involved no clinical interventions. The IRB ensured that appropriate data security and privacy measures were in place.

We include the full approval letter in the journal's manuscript portal. 

\section{Masking Raw Coordinates as a Privacy Benefit}
\label{appendix:privacy}

To demonstrate the potential benefit of our approach, we construct a hypothetical scenario where masking raw original coordinate can preserve individual privacy. Consider a city with a land area of $A = 1000\,\text{km}^2$ and a GPS sampling resolution of $r = 10\,\text{m}$. The number of unique spatial states $N_{spatial}$ is $A / r^2 = 10^7$, yielding a spatial entropy of:
\begin{equation}
H(location) = \log_2(N_{spatial}) = \log_2(10^7) \approx 26.57 \text{ bits.}
\end{equation}
In contrast, assuming a speed range of $0$--$120\,\text{km/h}$ quantized at $1\,\text{km/h}$ increments ($121$ states), the speed entropy is:
\begin{equation}
H(speed) = \log_2(N_{speed}) = \log_2(121) \approx 6.92 \text{ bits.}
\end{equation}
This indicates that a single speed entry provides less than one-third of the information necessary to resolve a spatial position. This reduction in entropy manifests in much higher $k$-anonymity \citep{sweeny2002}: while four spatio-temporal coordinates can uniquely identify 95\% of individuals ($k=1$) \citep{de2013unique}, speed sequences in dense urban environments can be far less distinctive. Numerous users share near-identical speed profiles due to shared traffic constraints such as synchronized signal timings and speed limits, making individual re-identification from speed sequences orders of magnitude more difficult than from absolute trajectories.

\section{Temporal Sampling Frequency}
\label{appendix:temporal}

\begin{figure}[ht]
    \centering
    \includegraphics[width=1\textwidth]{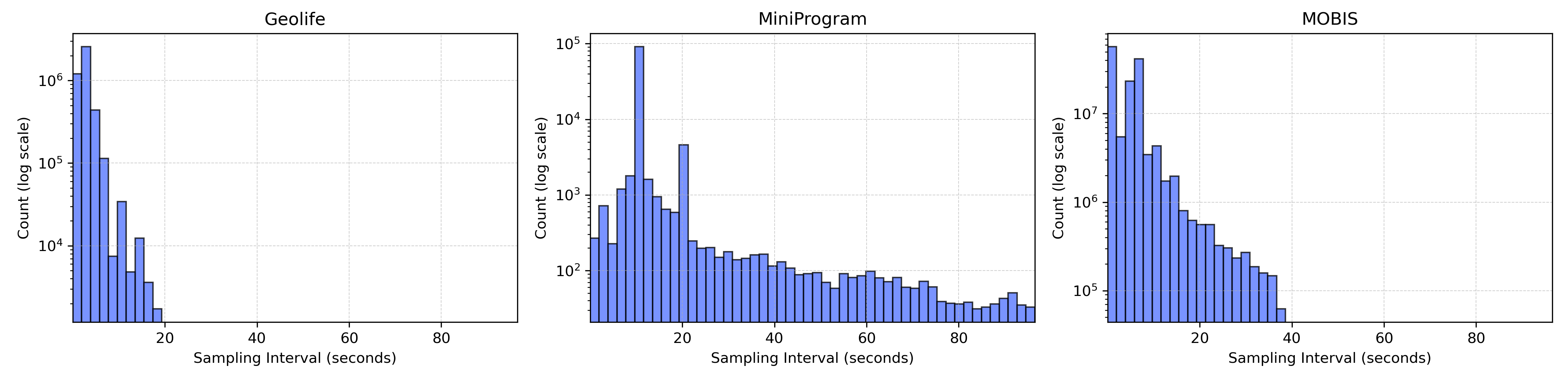}
    \caption{Sampling Frequency Distribution}
    \label{fig:sampling_frequency}
\end{figure}

We use dense GPS trajectories in this study, primarily obtained through continuous app-based tracking on personal smartphones and mobile devices. This stands in contrast to sparse GPS trajectories derived from geolocated social media posts or sporadic geotagged check-in data. As such, the temporal sampling frequency of our dense GPS data is generally high, though it still varies across datasets. For all three datasets used in this study, we retained the original sampling rates and did not perform any temporal resampling. Figure~\ref{fig:sampling_frequency} shows the distribution of time intervals between consecutive GPS samples in each dataset. By comparing trajectories collected through our \textit{CarbonClever} mini-program field experiment with those from Geolife and MOBIS, we observe variations in temporal granularity. These variations provide a rigorous basis for evaluating model performance. 

\section{Architectural Comparison of Transformer-based Models}
\label{appendix:parameters}

The number of trainable parameters in a Transformer-based model is important because it reflects a fundamental trade-off between model performance and computational demand. In general, models with more parameters require greater computational resources and larger training datasets to achieve strong performance. The numbers of trainable parameters for SpeedTransformer and the related Transformer variants are summarized in Table~\ref{tab:arch_comparison}. As shown, SpeedTransformer does not contain substantially more parameters than other state-of-the-art Transformer-based mode-detection models.

\begin{table}[ht]
\centering
\footnotesize
\setlength{\tabcolsep}{5pt} 
\caption{Architectural Comparison of Transformer-based Mobility Models}
\label{tab:arch_comparison}
\begin{tabular}{lccccc}
\toprule
\textbf{Model} & \textbf{Pre-proc.} & \textbf{Input} & \textbf{Attention} & \textbf{PosEnc.} & \textbf{Params} \\
\midrule
\textbf{SpeedTransformer} & Speed Calc. & 1D Speed & GQA & RoPE & 0.73M \\ 
\addlinespace
Deep-ViT \citep{ribeiro_deep_2024} & Image Gen. & 2D Image & Vision MHA & Learnable & 0.53M \\ 
\addlinespace
TMD-BERT \citep{drosouli2023tmd} & Discretisation & Tokens & Bi-Directional MHA & Learnable & 110M \\ 
\bottomrule
\multicolumn{6}{p{\textwidth}}{\scriptsize Note: GQA: Grouped-Query Attention; MHA: Multi-Head Attention; RoPE: Rotary Positional Embeddings; Image Gen.: DeepInsight Transform; Params: approximate number of trainable parameters using the best configuration.}
\end{tabular}
\end{table}

\section{Confusion Matrices and Class Imbalance}
\label{appendix:confusion_matrices}

\begin{figure*}[t]
\centering
\begin{tabular}{@{}c@{\hspace{10pt}}c@{}}
\includegraphics[width=0.45\textwidth]{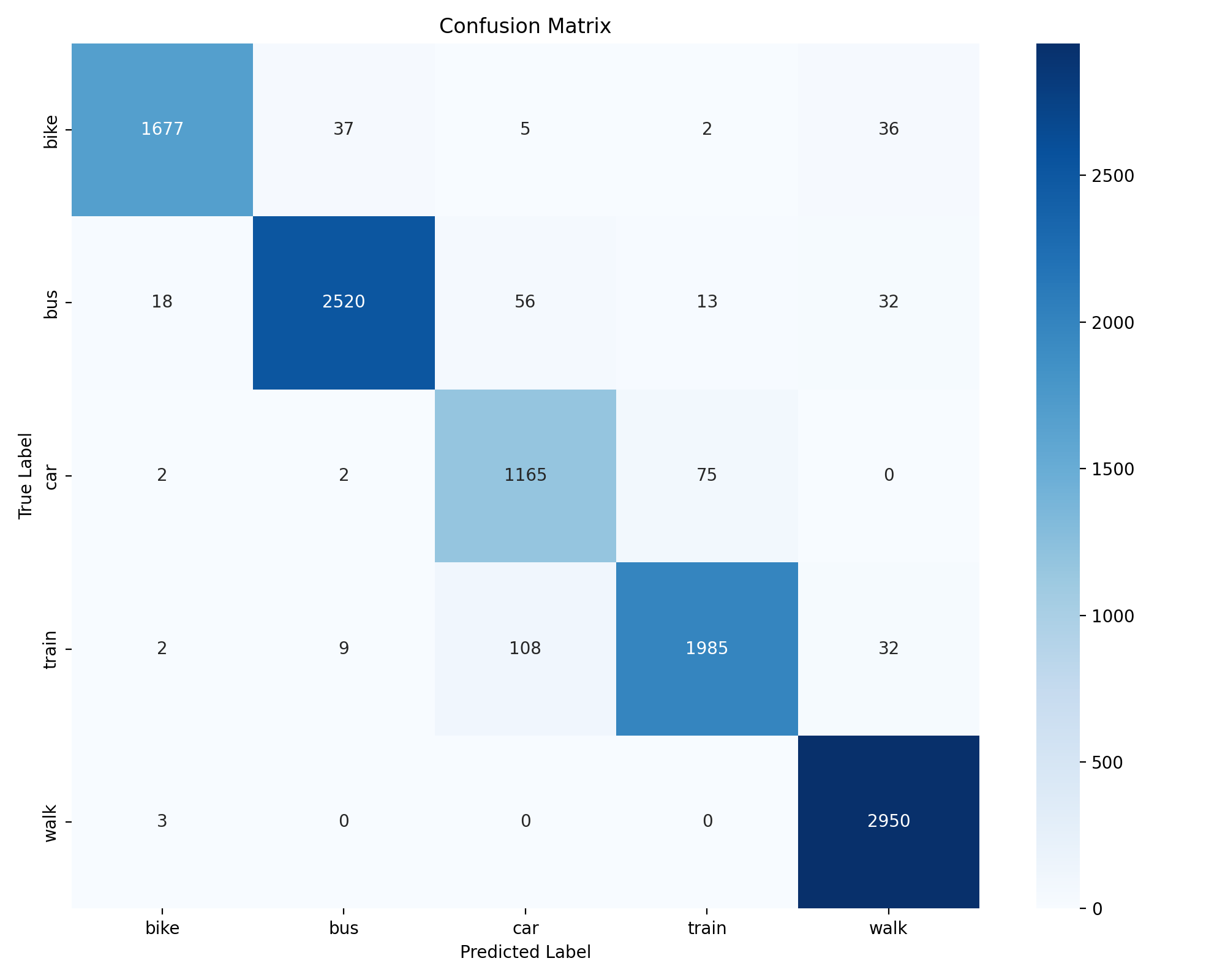} &
\includegraphics[width=0.45\textwidth]{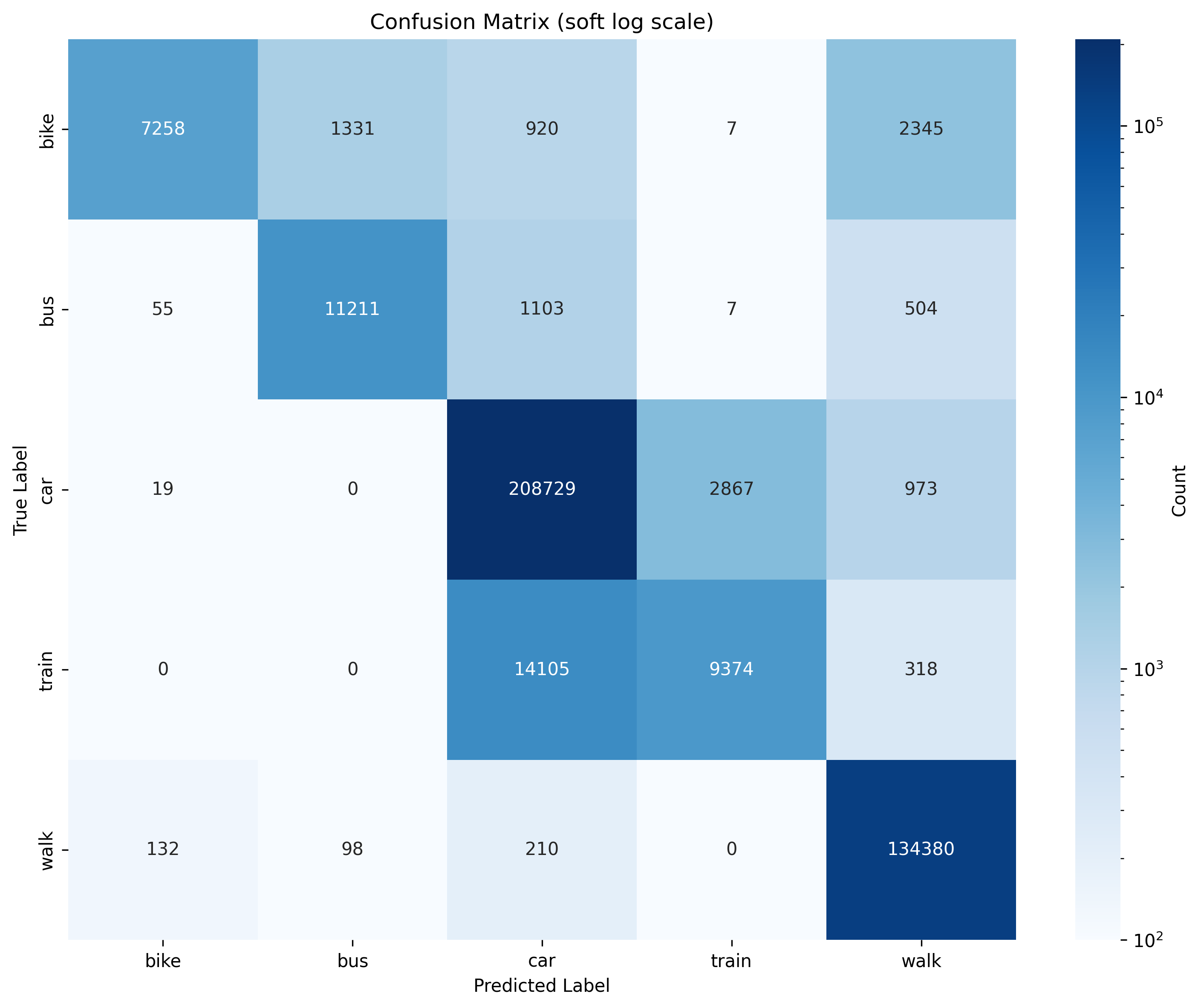} \\
(a) Geolife & (b) MOBIS
\end{tabular}
\caption{Confusion matrices for SpeedTransformer on the Geolife and MOBIS test sets. Values are raw counts; MOBIS exhibits stronger class imbalance, leading to larger absolute counts.}
\label{fig:conf_mats_appendix}
\end{figure*}

We examined the extent to which class imbalance in the training data—specifically the substantially smaller number of observations for \textit{Bike}, \textit{Bus}, and \textit{Train} in both MOBIS and Geolife—contributes to misclassification. Figure~\ref{fig:conf_mats_appendix} reports the confusion matrices for SpeedTransformer on the Geolife and MOBIS test sets, respectively. As is typical across machine-learning algorithms, SpeedTransformer exhibits weaker performance on classes with limited training data, namely \textit{Bike}, \textit{Bus}, and \textit{Train}, relative to classes that are more abundantly represented. Nevertheless, the performance is far from abysmal: per-class accuracy remains well above 50\%.

As shown in Fig.~\ref{fig:conf_mats_appendix}, the Geolife dataset yields a clear, strongly diagonal confusion matrix, indicating consistent class separation. By contrast, in MOBIS, the \textit{Train} category effectively aggregates several rail- and public-transport subclasses (e.g., train, tram, and metro-style services), whose movement signatures are often highly similar. As a result, predictions for \textit{Train} more frequently diffuse into adjacent public-transport labels.

\begin{figure}[t]
\centering
\includegraphics[width=1\textwidth]{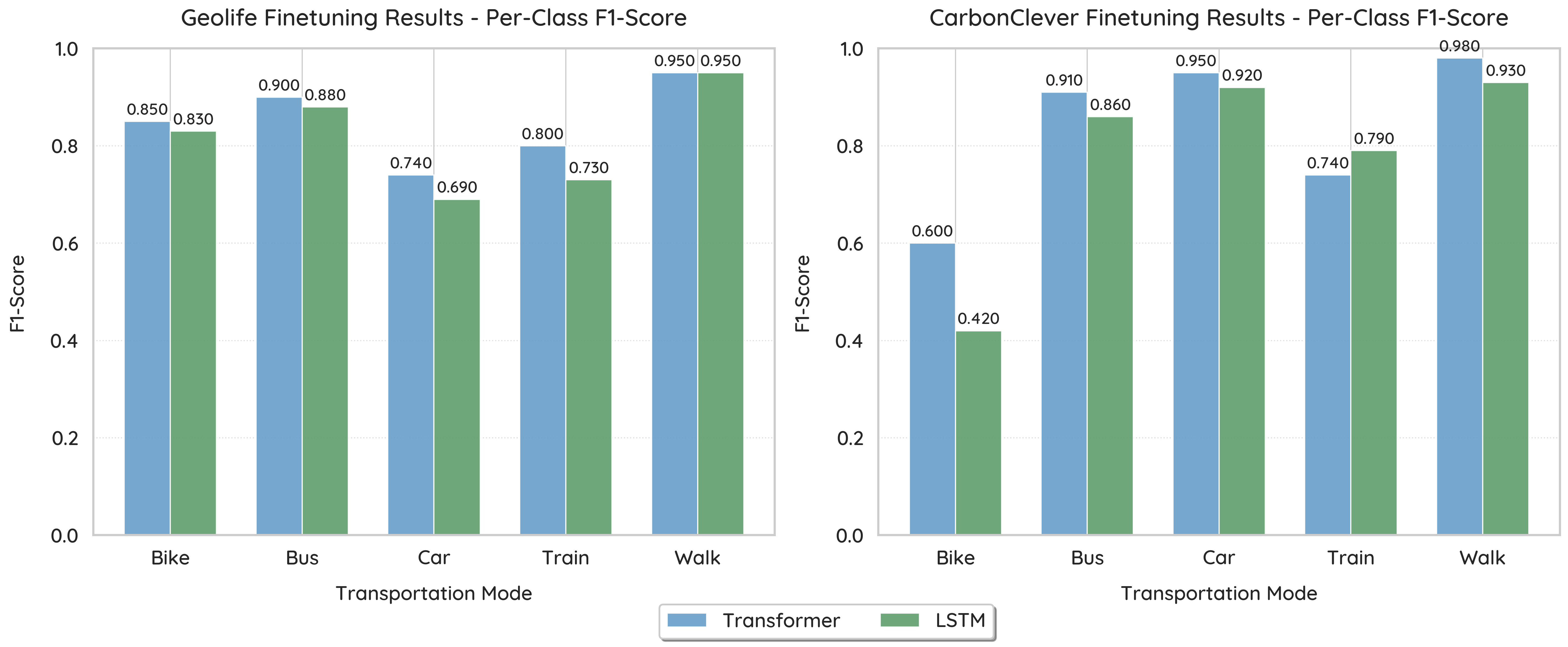}
\caption{Per class F1-Score for Fine-tuning on Geolife and our real-world field experiment data using SpeedTransformer and LSTM. The SpeedTransformer consistently achieves better results than the LSTM-Attention across all classes in both tasks.}
\label{fig:transfer}
\end{figure}

Moreover, when SpeedTransformer is first trained on MOBIS—the dataset with the largest volume of training samples—and subsequently fine-tuned using a small subset of Geolife and our CarbonClever field-experiment data, its performance remains comparatively strong (Figure~\ref{fig:transfer}). Per-class accuracy is generally above 0.9, and even in classes where MOBIS training data may be less well matched to Geolife or the field-experiment test data (e.g., the \textit{Train} class), accuracy still exceeds 0.7.

\section{Failure Analysis: Train Mode Case Studies}
\label{appendix:failure_analysis}

\begin{figure}[t]
    \centering
    \includegraphics[width=0.9\textwidth]{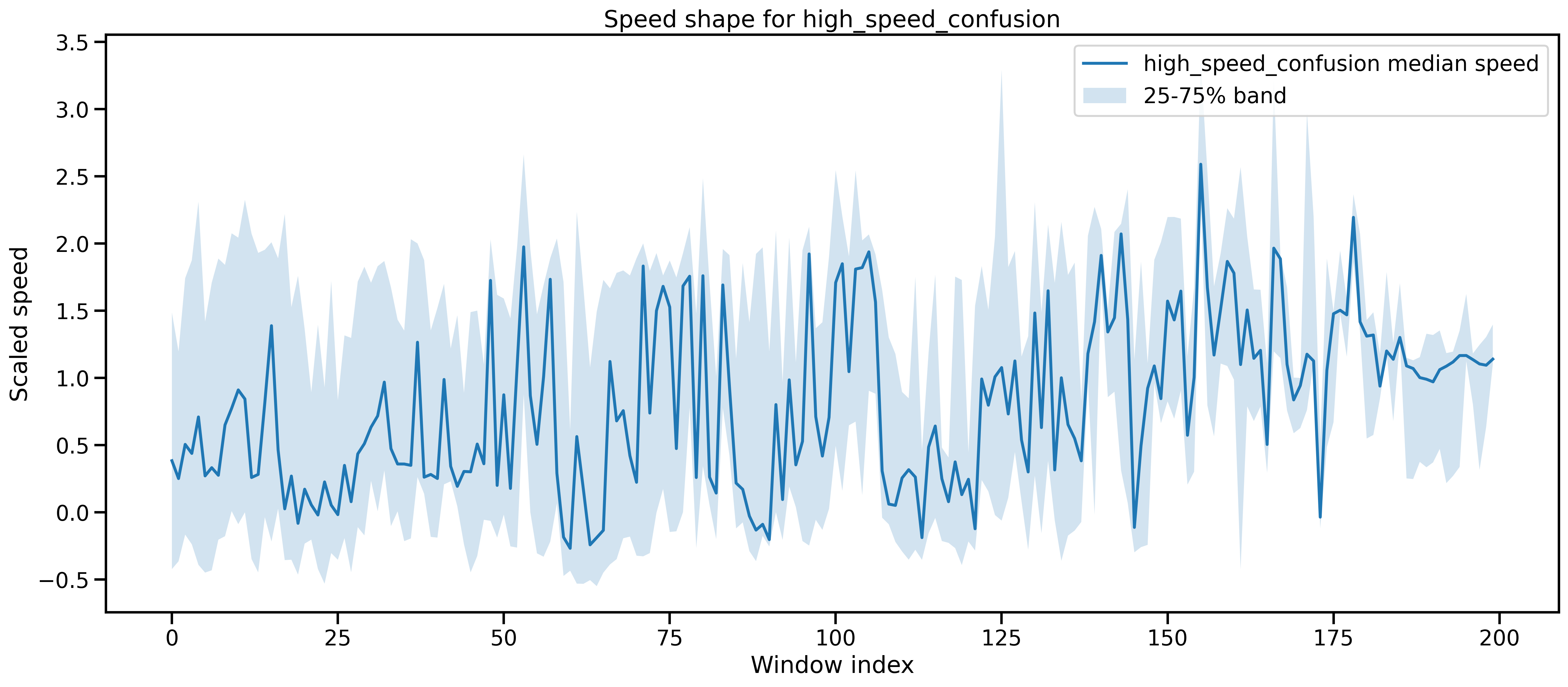}

    \vspace{0.5em}

    \includegraphics[width=0.9\textwidth]{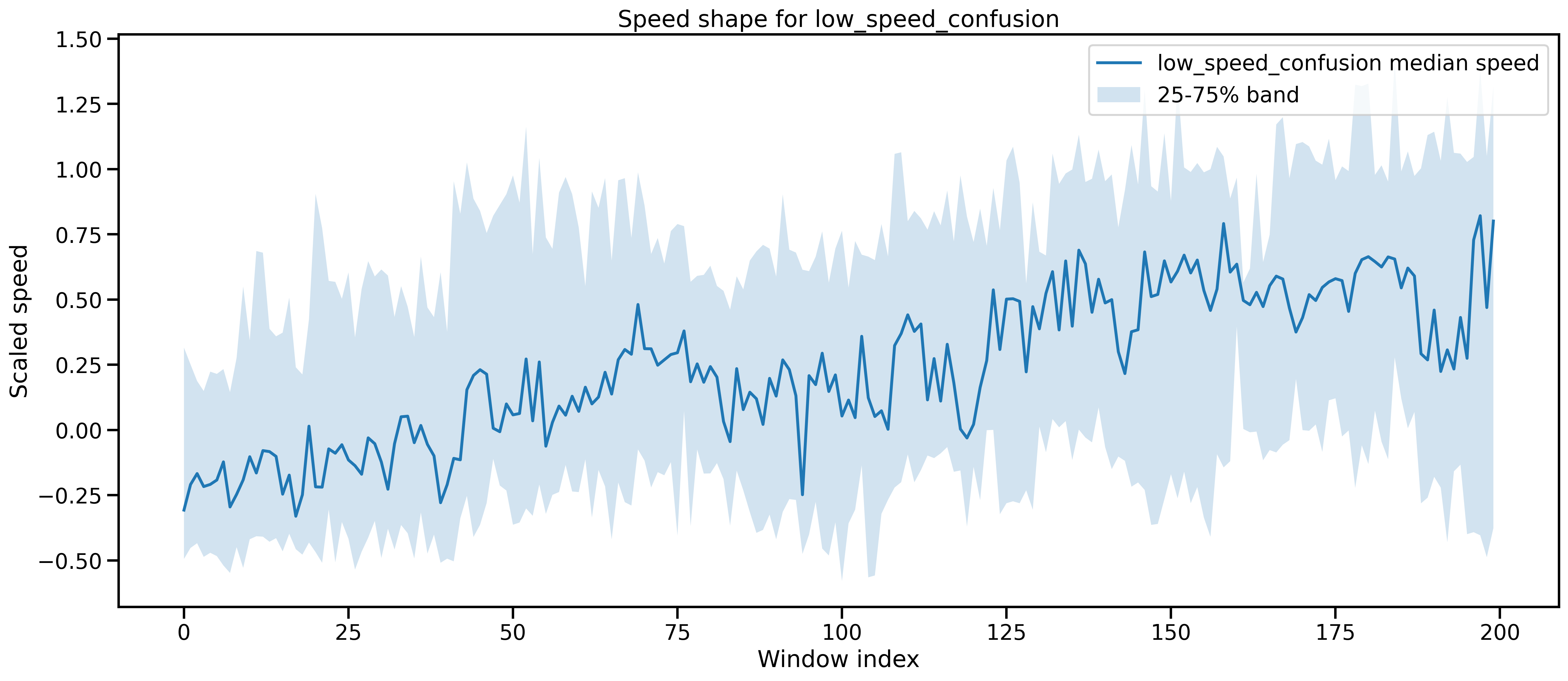}
    \caption{Sequence-level speed dynamics for two dominant train failure profiles. The \emph{low\_speed\_confusion} profile shows relatively smooth, gradual ramps; the \emph{high\_speed\_confusion} profile has higher variance with abrupt fluctuations and spikes. Shaded regions indicate the 25--75\% interquartile range across sequences.}
    \label{fig:speed_profiles_appendix}
\end{figure}

Having shown that misclassification is more prevalent for certain classes, we next examine specific failure patterns using the \textit{Train} mode as a case study. Figure~\ref{fig:speed_profiles_appendix} contrasts two dominant error profiles for train mode: a \emph{low\_speed\_confusion} group, which is frequently misclassified as walk or bus and exhibits smooth, gradually increasing speeds; and a \emph{high\_speed\_confusion} group, often involving car–train mix-ups, characterized by stop–go variability and intermittent speed spikes. Both profiles encompass many distinct trajectories, suggesting systematic patterns rather than isolated anomalies.

\begin{figure}[t]
    \centering
    \includegraphics[width=0.9\textwidth]{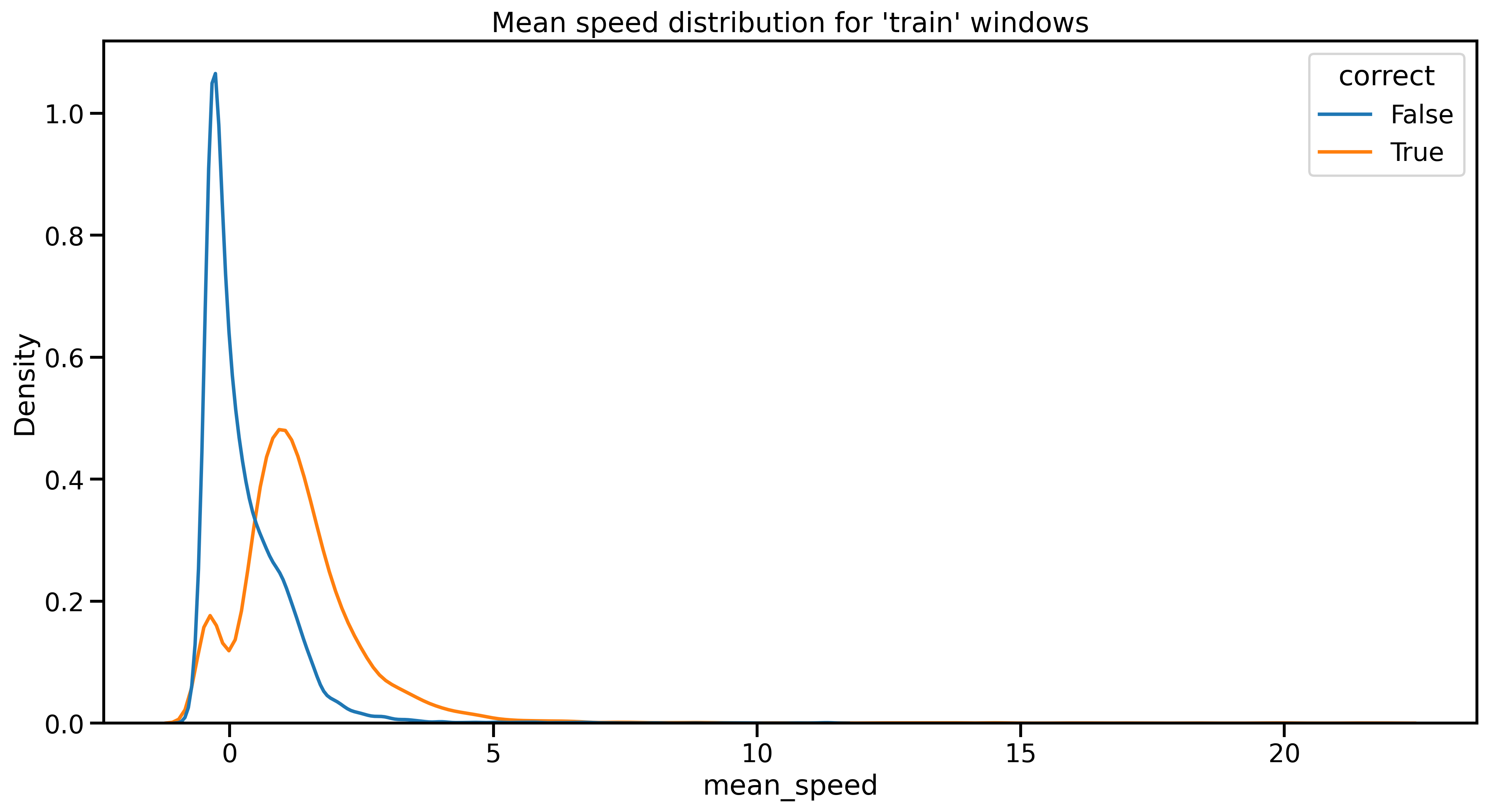}

    \vspace{0.5em}

    \includegraphics[width=0.9\textwidth]{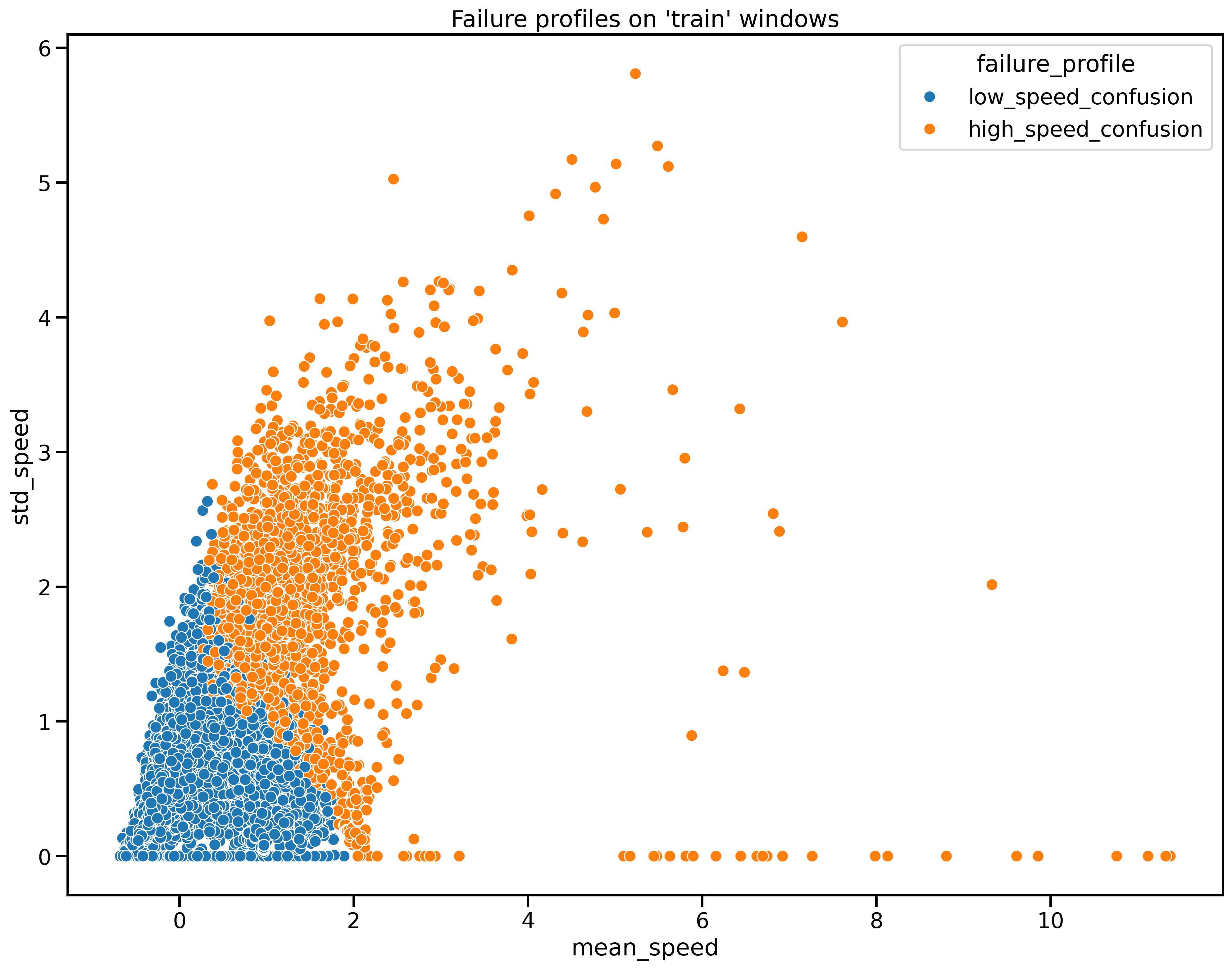}
    \caption{Distributional heterogeneity in train-class speed statistics. Mean speed and joint mean--std representations overlap substantially between correct and incorrect train windows, showing that simple aggregates cannot isolate the failure profiles (\emph{low\_speed\_confusion} vs \emph{high\_speed\_confusion}).}
    \label{fig:failure_stats_appendix}
\end{figure}

Figure~\ref{fig:failure_stats_appendix} further shows substantial overlap in simple summary statistics (e.g., mean and standard deviation) between correctly and incorrectly classified train mode, indicating that such aggregates are insufficient to distinguish the error. As such, the model learns a diffuse representation of the \textit{Train} class, which in turn allows leakage into neighboring public-transport modes. By contrast, Geolife exhibits more homogeneous train trajectories, supporting a sharper and more stable decision boundary for this class.

\end{document}